\RequirePackage{fix-cm}
\documentclass[twocolumn]{svjour3}
\smartqed
\usepackage{bm}
\usepackage{graphicx}
\usepackage{newtxtext, newtxmath}
\usepackage{natbib}
\usepackage{balance}
\usepackage{amsmath}
\usepackage[colorlinks=true,citecolor=blue]{hyperref}
\usepackage{xcolor}
\usepackage{latexsym}
\usepackage{makecell}
\usepackage{diagbox}
\usepackage{adjustbox}
\usepackage{caption}
\usepackage{subcaption}
\usepackage{soul}
\usepackage[ruled,linesnumbered]{algorithm2e}
\usepackage{booktabs}
\usepackage{multirow}
\usepackage{stfloats}
\usepackage{threeparttable}
\newcommand{\revise}[1]{\textcolor{black}{#1}}

\usepackage{hyperref}
\usepackage{url}
\usepackage{array}
\usepackage{ragged2e}
\usepackage{enumitem}
\usepackage{microtype}
\PassOptionsToPackage{hyphens}{url}\usepackage{hyperref}

\journalname{International Journal of Computer Vision}

\begin{document}

\title{GLENet: Boosting 3D Object Detectors with Generative Label Uncertainty Estimation}

\author{Yifan Zhang \and Qijian Zhang \and Zhiyu Zhu \and Junhui Hou \and Yixuan Yuan}

\institute{
Yifan Zhang, Qijian Zhang, Zhiyu Zhu, and Junhui Hou \at
Department of Computer Science, City University of Hong Kong. \\
\email{
\{yzhang3362-c, qijizhang3-c, zhiyuzhu2-c\}@my.cityu.edu.hk;
jh.hou@cityu.edu.hk; 
}
\and 
Yixuan Yuan \at
Department of Electronic Engineering, The Chinese University of Hong Kong. \\
\email{yxyuan@ee.cuhk.edu.hk}
\and 
%This project was supported by the Hong Kong Research Grants Council under Grants 11202320 and 11218121. 
This work was supported in part by the Hong Kong Research Grants Council under Grant 11202320 and Grant 11219422, and in part by the Hong Kong Innovation and Technology Fund under Grant MHP/117/21.
%and partly by the Natural Science Foundation of China under Grant 61871342. 
Corresponding author: Junhui Hou
}

\date{Received: date / Accepted: date}
% The correct dates will be entered by the editor

\maketitle
\begin{abstract}
The inherent ambiguity in ground-truth annotations of 3D bounding boxes, caused by occlusions, signal missing, or manual annotation errors, can confuse deep 3D object detectors during training, thus deteriorating detection accuracy. However, existing methods overlook such issues to some extent and treat the labels as deterministic. In this paper, we formulate the label uncertainty problem as the diversity of potentially plausible bounding boxes of objects. Then, we propose GLENet, a generative framework adapted from conditional variational autoencoders, to model the one-to-many relationship between a typical 3D object and its potential ground-truth bounding boxes with latent variables. The label uncertainty generated by GLENet is a plug-and-play module and can be conveniently integrated into existing deep 3D detectors to build probabilistic detectors and supervise the learning of the localization uncertainty. Besides, we propose an uncertainty-aware quality estimator architecture in probabilistic detectors to guide the training of the IoU-branch with predicted localization uncertainty. We incorporate the proposed methods into various popular base 3D detectors and demonstrate significant and consistent performance gains on both KITTI and Waymo benchmark datasets. Especially, the proposed GLENet-VR outperforms all published LiDAR-based approaches by a large margin and achieves the top rank among single-modal methods on the challenging KITTI test set.
\revise{The source code and pre-trained models are publicly available at \url{https://github.com/Eaphan/GLENet}.}

\end{abstract}

\section{Introduction} \label{sec1}

As one of the most practical application scenarios of computer vision, 3D object detection has been attracting much academic and industrial attention in the current deep learning era with the rise of autonomous driving and the emergence of large-scale annotated datasets (e.g., KITTI \citep{Geiger_KITTI}, and Waymo \citep{Sun_2020_CVPR}).

\begin{figure*}[htp]
\centering
\includegraphics[width=0.7\textwidth]{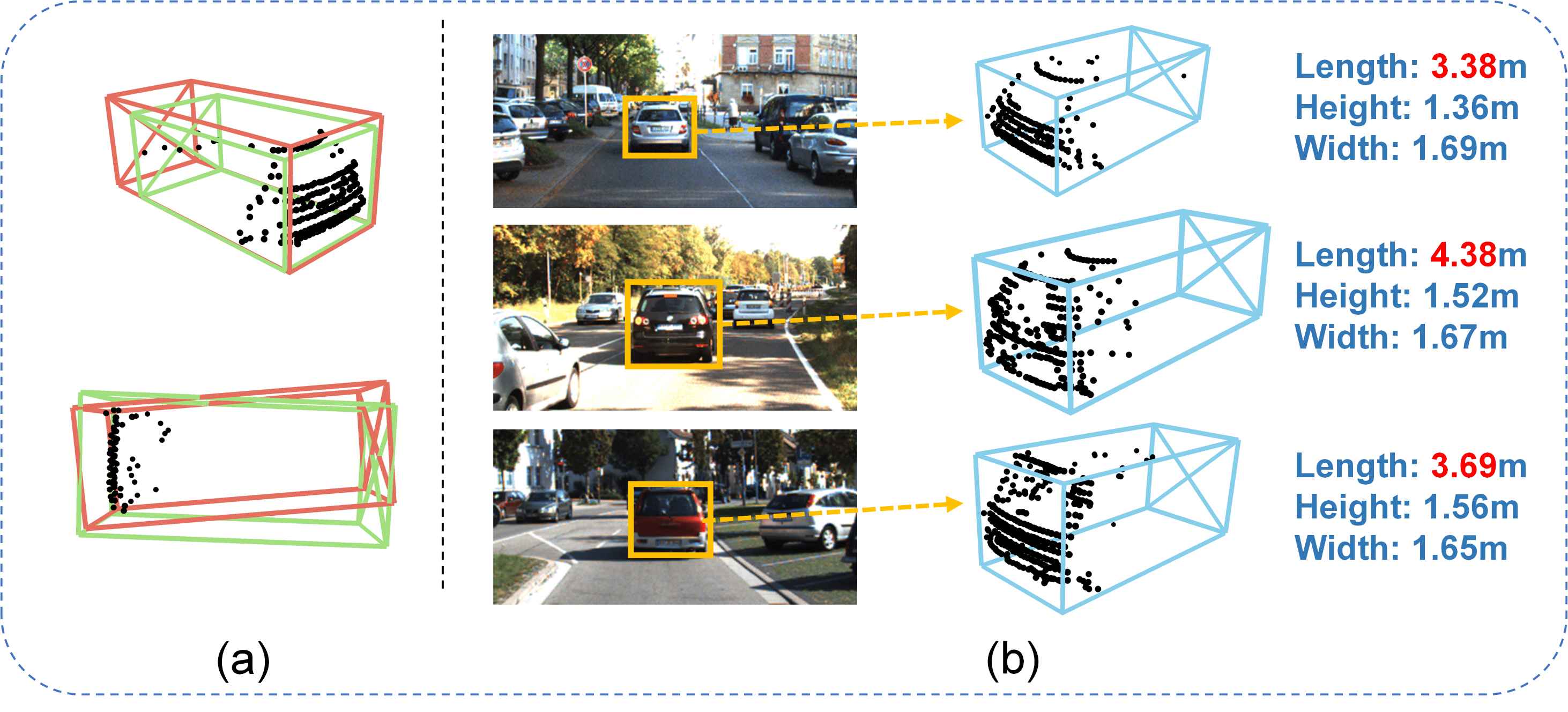}
\caption{(a) Given an object with an incomplete LiDAR observation, there may exist multiple potentially plausible ground-truth bounding boxes with varying sizes and shapes. (b) Ambiguity and inaccuracy can be inevitable in the labeling process when annotations are derived from 2D images and partial points. In the given cases, similar point clouds of the \textit{car} category with only the \textit{rear} part can be annotated with different ground-truth boxes of varying lengths.}
\label{fig:example}
\end{figure*}

In the current community, despite the proliferation of various deep learning-based 3D detection pipelines, it is observed that mainstream 3D object detectors are typically designed as deterministic models, without considering the critical issue of the ambiguity of annotated ground-truth labels. However, different aspects of ambiguity/inaccuracy inevitably exist in the ground-truth annotations of object-level bounding boxes, which may significantly influence the overall learning process of such deterministic detectors. For example, in the data collection phase, raw point clouds can be highly incomplete due to the intrinsic properties of LiDAR sensors as well as uncontrollable environmental occlusion. Moreover, in the data labeling phase, ambiguity naturally occurs when different human annotators subjectively estimate object shapes and locations from 2D images and partial 3D points. To facilitate intuitive understandings, we provide typical examples in Fig.~\ref{fig:example}, from which we can observe that an incomplete LiDAR observation can correspond to multiple potentially plausible labels, and objects with similar LiDAR observation can be annotated with significantly varying bounding boxes.

\begin{figure*}
\centering
\includegraphics[width=0.8\textwidth]{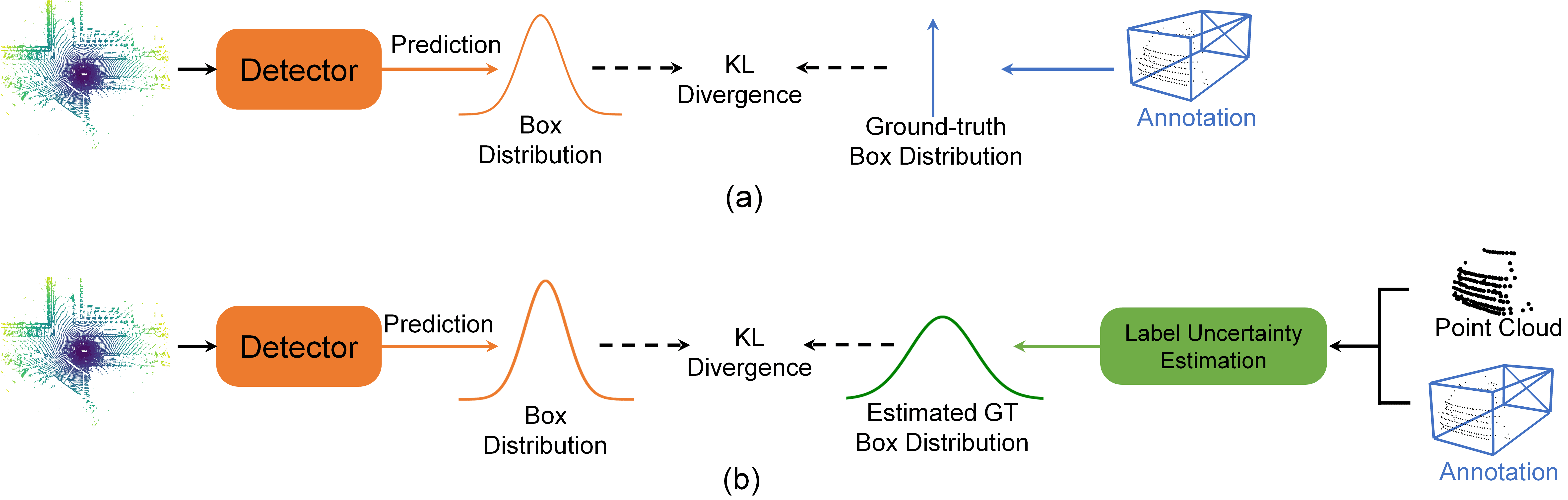}
\caption{Illustration of two different learning paradigms of probabilistic object detectors. (a) Methods that adopt probabilistic modeling in the detection head but essentially still ignore the issue of ambiguity in ground-truth bounding boxes. (b) Methods that explicitly estimate ground-truth bounding box distributions to be used as more reliable supervision signals.}
\label{fig:pod_with_lu}
\end{figure*}

Motivated by the aforementioned phenomena, there also exists another family of probabilistic detectors that explicitly consider the potential influence of label ambiguity. Conclusively, these methods can be categorized into two paradigms, as illustrated in Fig.~\ref{fig:pod_with_lu}. The first paradigm of learning frameworks \citep{bbr,LaserNet,f1,f2} tends to output the probabilistic distribution of bounding boxes instead of directly regressing definite box coordinates in a deterministic fashion. For example, under the pre-assumption of a Gaussian distribution, the detection head predicts the mean and variance of the distribution accordingly. To supervise such probabilistic models, these works simply treat ground-truth bounding boxes as the Dirac delta distribution, after which KL divergence is applied between the estimated distributions and ground truths. Obviously, the major limitation of these methods lies in that they fail to essentially address the problem of label ambiguity, since the ground-truth bounding boxes are still considered deterministic with zero uncertainty (i.e., modeled as a Dirac delta function). To this end, the second paradigm of learning frameworks attempts to quantify label uncertainty derived from some simple heuristics (\cite{meyer2020learning}) or Bayes (\cite{feng_iros}), such that the detectors can be supervised under a more reliable bounding box distribution. However, it is unsurprising that these approaches still cannot produce satisfactory label uncertainty estimation results due to insufficient modeling capacity. In general, this line of work is still in its initial stage with a very limited number of studies, despite its greater potential in generating higher-quality label uncertainty estimation in a data-driven manner.

\begin{figure*}[htp]
\centering
\includegraphics[width=\textwidth]{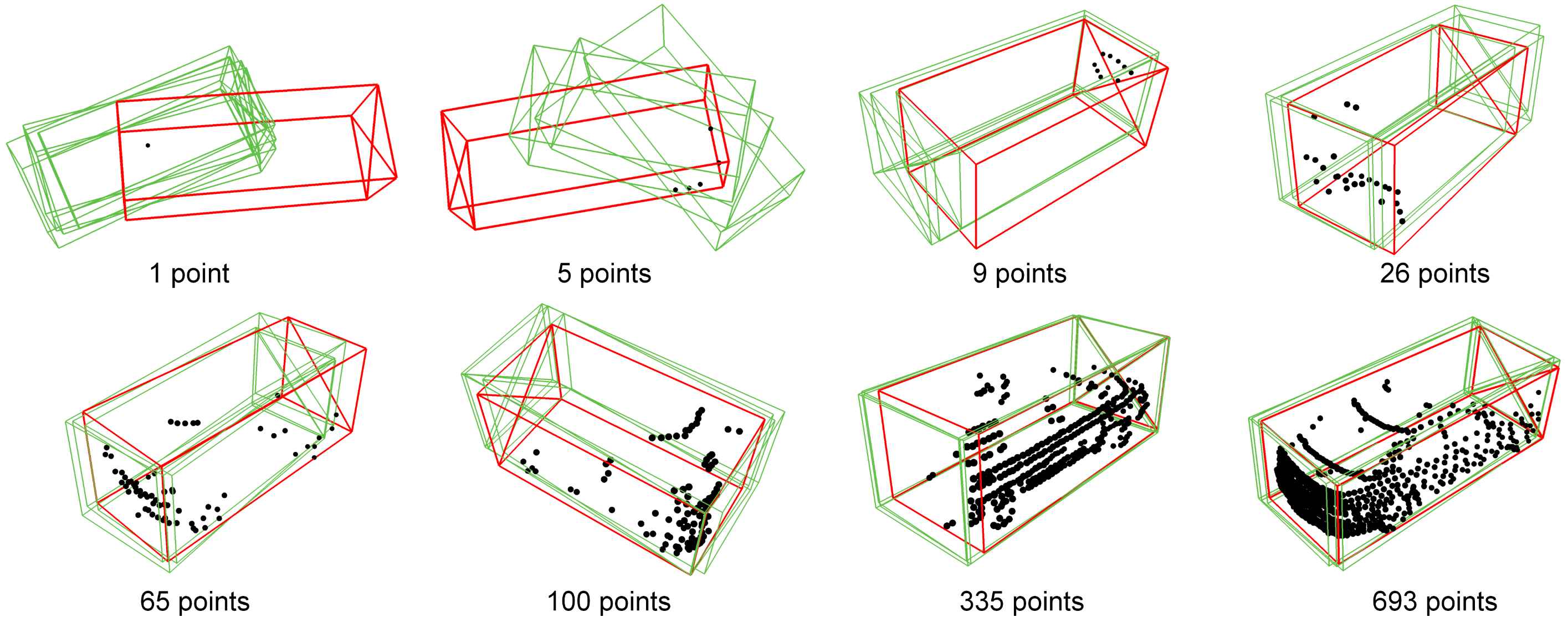}
\caption{Illustration of multiple potentially plausible bounding boxes from GLENet \revise{on the KITTI dataset} by sampling \revise{latent} variables multiple times. The point cloud, annotated ground-truth boxes, and predictions of GLENet are colored in black, red, and green, respectively. GLENet produces diverse predictions for objects represented with sparse point clouds and incomplete outlines, and consistent bounding boxes for objects with high-quality point clouds. The variance of the multiple predictions by GLENet is used to estimate the uncertainty of the annotated ground-truth bounding boxes.}
\label{fig:plenet_out}
\end{figure*}

Architecturally, this work follows the second type of design philosophy, where we particularly customize a powerful deep learning-based label uncertainty quantification framework to enhance the reliability of the estimated ground-truth bounding box distributions. Technically, we formulate the label uncertainty problem as the diversity of potentially plausible bounding boxes and explicitly model the one-to-many relationship between a typical 3D object and its potentially plausible ground-truth boxes in a learning-based framework. We propose GLENet, a novel deep generative network adapted from conditional variational auto-encoders (CVAE), which introduces a latent variable to capture the distribution over potentially plausible bounding boxes of point cloud objects. During inference, we sample latent variables multiple times to generate diverse bounding boxes (see Fig.\ref{fig:plenet_out}), the variance of which is taken as label uncertainty to guide the learning of localization uncertainty estimation in the downstream detection task. Besides, based on the observation that detection results with low localization uncertainty in probabilistic detectors tend to have accurate actual localization quality (see Section\ref{sec:UAQE}), we further propose the uncertainty-aware quality estimator (UAQE), which facilitates the training of the IoU-branch with the localization uncertainty estimation.

To demonstrate our effectiveness and universality, we integrate GLENet into several popular 3D object detection frameworks to build powerful probabilistic detectors. Experiments on KITTI \citep{Geiger_KITTI} and Waymo \citep{Sun_2020_CVPR} datasets demonstrate that our method can bring consistent performance gains and achieve the current state-of-the-art. Particularly, the proposed GLENet-VR surpasses all published single-modal detection methods by a large margin and \textbf{ranks $1^{st}$} among all published LiDAR-based approaches on the highly competitive KITTI 3D detection benchmark on March 29$^{th}$, 2022\footnote{
% \url{}
\href{www.cvlibs.net/datasets/kitti/eval_object.php?obj_benchmark=3d}{www.cvlibs.net/datasets/kitti/eval\_object.php?obj\_benchmark=3d}
% \hyperref[www.cvlibs.net/datasets/kitti/eval_object.php?obj_benchmark=3d]{www.cvlibs.net/datasets/kitti/eval\_object.php?obj\_benchmark=3d}
}. 

We summarize the main contributions of this paper as follows:

\begin{itemize}
\item We are the first to formulate the 3D label uncertainty problem as the diversity of potentially plausible bounding boxes of objects. To capture the one-to-many relationship between a typical 3D object and its potentially plausible ground-truth bounding boxes, we present a deep generative model named GLENet. Additionally, we introduce a general and unified deep learning-based paradigm, including the network structure, loss function, evaluation metric, etc.
\item Inspired by the strong correlation between the localization quality and the predicted uncertainty in probabilistic detectors, we propose UAQE to facilitate the training of the IoU-branch.
\end{itemize}

The remainder of the paper is organized as follows. 
Section \ref{sec:RW} reviews existing works on 
%\JHdel{shows the related work including reviews} 
 LiDAR-based detectors and label uncertainty estimation methods. 
% \textcolor{blue}{Section~\ref{sec: proposed method} describes our architecture and how to estimate and leverage the label uncertainty.}
In Section~\ref{sec: proposed method}, we explicitly formulate the label uncertainty estimation problem from the probabilistic distribution perspective, followed by the technical implementation of GLENet.
In Section \ref{bbr_label_uncertainty}, we introduce a unified way of integrating the label uncertainty statistics predicted by GLENet into the existing 3D object detection frameworks to build more powerful probabilistic detectors, as well as some theoretical analysis.
In Section \ref{experiments}, we conduct experiments on the KITTI dataset and the Waymo Open dataset to demonstrate the effectiveness of our method in enhancing existing 3D detectors and the ablation study to analyze the effect of different components. Finally, Section \ref{sec:con} concludes this paper.

% \JHNOTE{merge this paragraph to the end of introduction} In what follows, we will explicitly formulate the label uncertainty estimation problem from the probabilistic distribution perspective,  %\st{provide the problem formulation} followed by the technical implementation of GLENet in Section \ref{method_GLENet}.

\section{Related Work}
\label{sec:RW}
\subsection{LiDAR-based 3D Object Detection}
Existing 3D object detectors can be classified into two categories: single-stage and two-stage. For single-stage detectors, \cite{Zhou_2018_CVPR} proposed to convert raw point clouds to regular volumetric representations and adopted voxel-based feature encoding. \cite{yan2018second} presented a more efficient sparse convolution. \cite{Lang_2019_CVPR} converted point clouds to sparse fake images using pillars. \cite{Shi_2020_CVPR} aggregated point information via a graph structure. \cite{He_2020_CVPR} introduced point segmentation and center estimation as auxiliary tasks in the training phase to enhance model capacity. %Zheng et al.
\cite{zheng2021cia} constructed an SSFA module for robust feature extraction and a multi-task head for confidence rectification, and proposed DI-NMS for post-processing. For two-stage detectors, %Shi et al.
\cite{shi2020points} exploited a voxel-based network to learn the additional spatial relationship between intra-object parts under the supervision of 3D box annotations. %Shi et al.
\cite{shi2019pointrcnn} proposed to directly generate 3D proposals from raw point clouds in a bottom-up manner, using semantic segmentation to validate points to regress detection boxes. The follow-up work \citep{yang2019std} further proposed PointsPool to convert sparse proposal features to compact representations and used spherical anchors to generate accurate proposals. \cite{shi2020pv} utilized both point-based and voxel-based methods to fuse multi-scale voxel and point features. \cite{deng2021voxel} proposed voxel RoI pooling to extract RoI features from coarse voxels.

To address the boundary ambiguity problems in 3D object detection caused by occlusion and signal miss, some studies, such as SPG \citep{Xu_2021_ICCV}, have tried to use point cloud completion methods to restore the full shape of objects and improve the detection performance \citep{yan2021sparse,Najibi_2020_CVPR}. However, generating complete and precise shapes with incomplete point clouds remains a non-trivial task.

\subsection{Probabilistic 3D Object Detector}
There are two types of uncertainty in deep learning predictions. A type of uncertainty called aleatoric uncertainty is caused by the inherent noise in observational data, which cannot be eliminated. The other type is called epistemic Uncertainty or model uncertainty, which is caused by incomplete training and can be alleviated with more training data.
Most existing state-of-the-art 2D~\citep{liu2016ssd,tan2020efficientdet,carion2020end} and 3D~\citep{shi2020points} object detectors produce a deterministic box with a confidence score for each detection. While the probability score represents the existence and semantic confidence, it cannot reflect the uncertainty about predicted localization well.
By contrast, probabilistic object detectors \citep{bbr,harakeh2020bayesod,varamesh2020mixture} estimate the probabilistic distribution of predicted bounding boxes rather than take them as deterministic results. For example, \cite{bbr} and \cite{Choi_2019_ICCV} modeled the predicted boxes as Gaussian distributions, the variance of which can indicate the localization uncertainty and is predicted with additional layers in the detection head. It introduces the KL Loss between the predicted Gaussian distribution and the ground-truth bounding boxes modeled as a Dirac delta function, so the regression branch is expected to output a larger variance and get a smaller loss for inaccurate localization estimation for the cases with ambiguous boundaries. 
% Unlike the common practice of modeling the box as a Gaussian distribution, \cite{harakeh2020bayesod} learned the off-diagonal elements of the covariance matrix of a multivariate Gaussian distribution as uncertainty estimation. 
\cite{li2021generalized} facilitated the learning of localization quality with distribution statistics of a bounding box, such as the mean value, which inspires us to further utilize the estimated uncertainty in UAQE.
\cite{LaserNet} proposed a probabilistic 3D object detector modeling the distribution of bounding box corners as a Laplacian distribution.

However, most probabilistic detectors take the ground-truth bounding box as a deterministic Dirac delta distribution and ignore the ambiguity in the ground-truth. 
Therefore, the localization variance is actually learned in an unsupervised manner, which may result in sub-optimal localization precision and erratic training (see our theoretical analysis in Section \ref{sec:theoretical analysis}).

\subsection{Label Uncertainty Estimation}
Label noise (or uncertainty) is a common problem in real-world datasets and could seriously affect the performance of supervised learning algorithms. 
As the neural network is prone to overfit to even complete random noise (\cite{zhang2021understanding}), it is important to prevent the network from overfitting noisy labels.
An obvious solution is to consider the label of a misclassified sample to be uncertain and remove the samples \citep{delany2012profiling}.
\cite{garcia2015using} used a soft voting approach to approximate a noise level for each sample based on the aggregation of the noise degree prediction calculated for a set of binary classifiers. \cite{luengo2018cnc} extended this work by correcting the label when most classifiers predict the same label for noisy samples.
Confident Learning~\cite{northcutt2021confident} estimated uncertainty in dataset labels by estimating the joint distribution of noisy labels and true labels.
However, the above studies mainly focus on the image classification task.

There only exists a limited number of previous works focusing on quantifying uncertainty statistics of annotated ground-truth bounding boxes. \cite{meyer2020learning} proposed to model label uncertainty by the IoU between the label bounding box and the corresponding convex hull of the aggregated LiDAR observations. 
However, it is non-learning-based and thus has limited modeling capacity. Besides, it only produces uncertainty of the ground-truth box as a whole instead of each dimension.
\cite{feng_iros} proposed a Bayes method to estimate label noises by quantifying the matching degree of point clouds for the given boundary box with the Gaussian Mixture Model. However, its assumption of conditional probabilistic independence between point clouds is often untenable in practice. Differently, we formulate label uncertainty as the diversity of potentially plausible bounding boxes. There may be some objects with few points that exactly match the learned surface points of the corresponding labeled bounding box, so the label is considered by \cite{feng_iros} to be deterministic. But for an object with sparse point clouds, our GLENet will output different and plausible bounding boxes and further estimate high label uncertainty based on them, regardless of whether points match the given label. In general, \cite{feng_iros} used the Bayesian paradigm to estimate the correctness of the annotated box as the label uncertainty, while our method formulates it as the diversity of potentially plausible bounding boxes and predicts it by GLENet.

\subsection{Conditional Variational Auto-Encoder}
The variational auto-encoder (VAE)~\citep{kingma2013auto} has been widely used in image and shape generation tasks~\citep{yan2016attribute2image,nash2017shape}. 
It transforms natural samples into a distribution where latent variables can be drawn and passed to a decoder network to generate diverse samples.
\revise{\cite{sohn2015learning} introduced the conditional variational autoencoder (CVAE), which extends the capabilities of the traditional VAE by incorporating an additional condition during the generative process. The CVAE model consists of an encoder, a decoder, and an extra input, which is usually a label or other structured information pertinent to the generation task. This auxiliary condition enables the CVAE to generate more targeted and controlled samples compared to its unsupervised counterpart, the VAE.}
In the NLP field, VAE has been widely applied to many text generation tasks, such as dialogue response~\citep{zhao2017learning}, machine translation~\citep{zhang-etal-2016-variational-neural}, story generation~\citep{wang2019t}, and poem composing~\citep{li2018generating}. VAE and CVAE have also been applied in computer vision tasks, like image generation~\citep{yan2016attribute2image}, human pose estimation~\citep{Sharma_2019_ICCV}, medical image segmentation~\citep{painchaud2020cardiac}, salient object detection~\citep{li2019supervae,zhang2020uc}, and modeling human motion dynamics~\citep{yan2018mt}.
Recently, VAE and CVAE algorithms have also been applied extensively to applications of 3D point clouds, such as generating grasp poses~\citep{mousavian20196} and instance segmentation~\citep{Yi_2019_CVPR}. 

Inspired by CVAE for generating diverse reasonable responses in dialogue systems, we propose GLENet adapted from CVAE to capture the one-to-many relationship between objects with incomplete point clouds and the potentially plausible ground-truth bounding boxes.
To the best of our knowledge, we are the first to employ CVAE in 3D object detection to model label uncertainty.

\section{Proposed Label Uncertainty Estimation}
\label{sec: proposed method}

\begin{figure*}[htp]
\centering
\includegraphics[width=\textwidth]{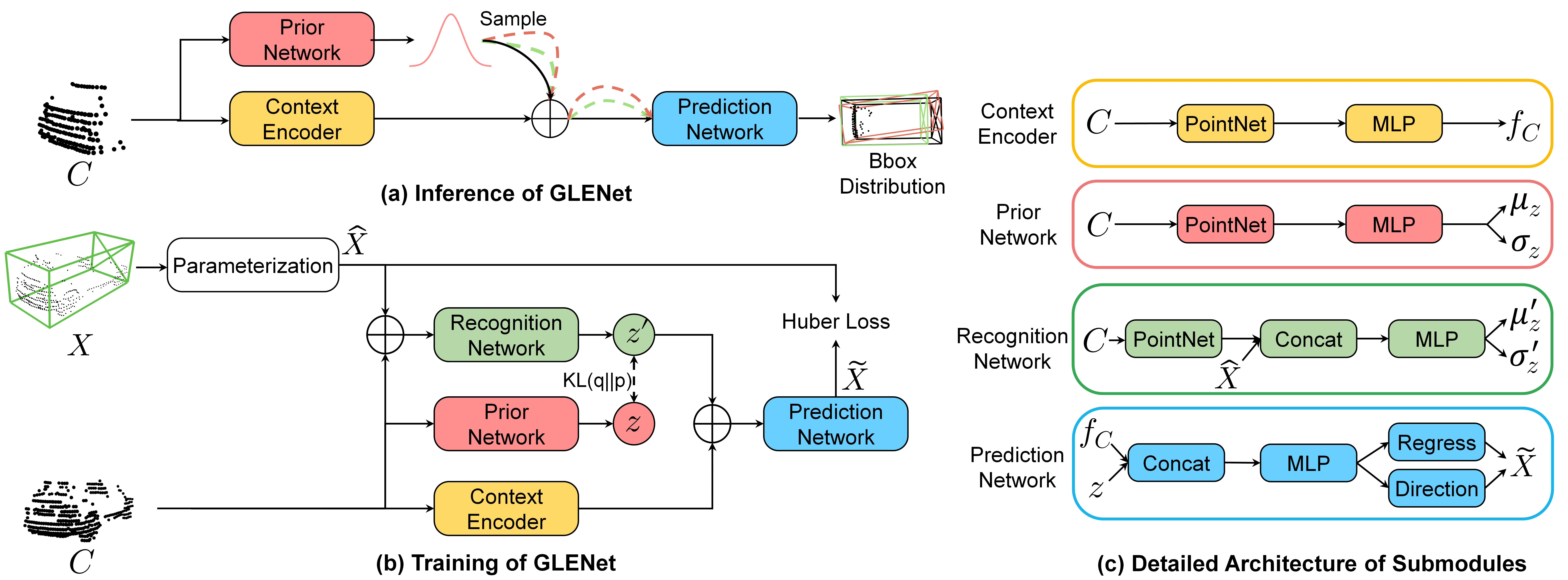} %\vspace{-0.3cm}
\caption{
The overall workflow of GLENet. In the training phase, we learn parameters $\mu$ and $\sigma$ (resp. $\mu^{\prime}$ and $\sigma^{\prime}$ ) of latent variable $z$ (resp. $z^{\prime}$) through the prior network (resp. recognition network), after which a sample of $z^{\prime}$ and the corresponding geometrical embedding produced by the context encoder are jointly exploited to estimate the bounding box distribution. In the %\textcolor{blue}{label} uncertainty estimation 
inference phase, we sample from the distribution of $z$ multiple times to generate different bounding boxes, whose variance we use as label uncertainty. \revise{Note we denote multiple sampling with black, orange, and green lines in subgraph (a).}
}
\label{fig:pipeline}
\end{figure*}

As aforementioned, the ambiguity of annotated ground-truth labels widely exists in 3D object detection scenarios and has adverse effects on the deep model learning process, which is not well addressed or even completely ignored by previous works. To this end, we propose GLENet, a generic and unified deep learning framework that generates label uncertainty by modeling the one-to-many relationship between point cloud objects and potentially plausible bounding box labels.
Then the variance of the multiple outputs of GLENet for a single object is computed as the label uncertainty, which is extended as an auxiliary regression objective to enhance the performance of the downstream 3D object detection task.
%, and extend it as an auxiliary regression objective to enhance 3D object detection performance.

%\setlength{\topskip}{0ex}

%\subsection{GLENet for Label Uncertainty Estimation} 
%\label{method_GLENet}
%\vspace{-0.1cm}

\subsection{Problem Formulation %\textcolor{blue}{\st{and Overview}}
} %of GLENet
Let $C=\{c_i\}^n_{i=1}$ be a set of $n$ observed LiDAR points belonging to an object, where $c_i \in \mathbb{R}^{3}$ is a 3D point represented with spatial coordinates. Let $X$ be the annotated ground-truth bounding box of $C$ parameterized by the center location $(c_x,c_y,c_z)$, the size ( length $l$ , width $w$, and height $h$), and the orientation $r$, i.e., $X=[c_x,~c_y,~c_z,~w,~l,~h,~r]\in \mathbb{R}^{7}$.

% \st{Inspired by previous works citep{bbr}, } 
% \YFNOTE{Considering the incompleteness of LiDAR observations and annotation ambiguity,} we assume the ground-truth bounding box subjects to a Gaussian distribution $q(X \vert C)$, whose expectation is exactly the value of the annotation and variance $\sigma^2$ indicates the degree of uncertainty about the annotation. The uncertainty can be approximated by the degree of confusion in the distribution of potential bounding boxes.
We formulate the uncertainty of the annotated ground-truth label of an object as the diversity of potentially plausible bounding boxes of the object, which could be quantitatively measured with the variance of the distribution of the potential bounding boxes.
First, we model 
the distribution of these potential boxes conditioned on point cloud $C$, denoted as $p(X \vert C)$.  
% To quantify label uncertainty, we propose to model the distribution of potentially plausible bounding boxes conditioned on point cloud $C$, i.e., $p(X \vert C)$, and take its variance as label uncertainty. 
% Considering the fact that directly modeling $p(X \vert C)$ can be intractable and result in inaccurate distribution estimation
%\textcolor{blue}{\st{It is inconvenient to use a discriminant model to estimate the distribution, i.e., to estimate $p(X \vert C)$ and then randomly generate $X$ to judge its probability.}}
%\JHNOTE{use a short sentence to explain the reason?},
Specifically, 
based on the Bayes theorem, we introduce an intermediate variable $z$ to write the conditional distribution as 
\begin{equation}
  p(X \vert C)=\int_{z}p(X \vert z,C)p(z \vert C)dz.
\end{equation}
% where $p(X \vert z,C)$ and $p(z \vert C)$ can be \textcolor{red}{deduced through neural networks parameterized by $\theta$}.
% \JHNOTE{As it is intractable to directly deduce the variance of $p(X \vert C)$ from the network,} 
Then, with $p(X \vert z,C)$ and $p(z\vert C)$ known, we can adopt a Monte \revise{Carlo} method to get multiple bounding box predictions by sampling $z$ multiple times and approximate the variance of $p(X \vert C)$ with that of the sampled predictions. %calculate the variance $\sigma^2$ of predictions to approximate %as approximation of
%the variance of $p(X \vert C)$ for each of the seven dimensions, i.e., $(c_x,c_y,c_z,w,l,h,r)$. 

In the following, we will introduce our learning-based framework named GLENet to realize the estimation process. 

% \noindent\textbf{Overall workflow of GLENet}.
\subsection{%\JHdel{Label Uncertainty Estimation Process} 
Inference Process of GLENet}
Fig.~\hyperref[fig:pipeline]{\ref{fig:pipeline}~(a)} shows %\JHdel{the label uncertainty estimation via} 
the flowchart of GLENet parameterized by neural parameters $\theta$, which aims to predict $p(z \vert C)$ and $p(X \vert z,C)$. 
%\JHdel{and we call the parameters involved in the inference process as $\theta$.} 
Specifically, under the assumption that the prior distribution $p(z \vert C)$ %$p_{\theta}(z \vert C)$ 
subjects to a multivariate Gaussian distribution parameterized by $(\mu_z,\sigma_z)$, denoted as $\mathcal{N}(\mu_{z}, \sigma_{z}^{2})$, we design a prior network, which is composed of PointNet~\citep{qi2017pointnet} and additional MLP layers, from the input point cloud $C$ to predict the values of $(\mu_z,\sigma_z)$. 
%\textcolor{blue}{The prior network is composed of PointNet~\citep{qi2017pointnet} and additional MLP layers.}
% \JHdel{where $(\mu_z,\sigma_z)$ denote vectorized parameters of the Gaussian distribution learned by the prior network.} 
% \JHdel{To model the bounding box distribution $p(X \vert z,C)$,}
%$p_{\theta}(X \vert z,C)$, 
Then, we employ a context encoder to embed the input point cloud $C$ into a high dimensional feature space, leading to the geometric feature representation $f_{C}$, which is concatenated with $z$ sampled from $\mathcal{N}(\mu_{z}, \sigma_{z}^{2})$ and fed into a prediction network composed of MLPs to regress the bounding box distribution $p(X \vert z,C)$, i.e., the localization, dimension, and orientation of the bounding box.   %\textcolor{blue}{\JHdel{As mentioned before, the variance of approximated $p(X \vert C)$ are taken as label uncertainty.}}

% \st{\noindent\textbf{Prior Network.} For the prior network, we adopt PointNet~\citep{qi2017pointnet} for point feature embedding and add additional MLP layers to regress $p_{\theta}(z \vert C)$, i.e., the values of ($\mu_z$, $\sigma_z$). \\}
As empirically observed in various related domains \citep{goyal2017z}, it could be difficult to make use of latent variables when the prediction network can generate a plausible output only using the sufficiently expressive features of condition $C$. Therefore, we utilize a simplified PointNet architecture as the backbone of the context encoder to avoid posterior collapse. 
\if 0
\noindent\textbf{Prediction Network.}
% Given $C$ and its bounding box $X$, we assume there is a true posterior distribution $p(z \vert X,C)$, and train the prediction network to restore $X$ from $z$ sampled from $p(z \vert X,C)$ and context features $f_{C}$. 
Given the estimated prior distribution $p_{\theta}(z \vert C)$, we utilize the prediction network to restore $X$ from $z$ sampled from $p(z \vert X,C)$ and context features $f_{C}$. 
\textcolor{blue}{$f_{C}$ and $z$ are concatenated and taken as the input of the following MLPs to predict localization, dimension, and orientation of the box.}
\fi 
We refer the readers to Section \ref{sec:imple details} for the implementation details of these modules. In the following sections, we also use $p_{\theta}(z \vert C)$, $p_{\theta}(X \vert z,C)$, and $p_{\theta}(X\vert C)$ to denote the predictions of $p(z \vert C)$, $p(X \vert z,C)$, and $p(X\vert C)$ by GLENet, respectively.

% As mentioned before, we obtain multiple predictions as the approximation of p(X \vert C) by sampling $z$ from distribution $p_{\theta}(z \vert C)$ multiple times
% \JHNOTE{Is it better to move this sentence to the end of the first paragraph of Section 3.2?}

\subsection{Training Process of GLENet}
\subsubsection{Recognition Network}
Given $C$ and its annotated bounding box $X$, we assume there is a true posterior distribution $q(z \vert X,C)$. %\JHdel{In order to make the prior distribution $p_{\theta}(z \vert C)$ close to $q(z \vert X,~C)$,}
Thus, during training, we construct a recognition network parameterized by network parameters $\phi$ (see Fig.~\hyperref[fig:pipeline]{\ref{fig:pipeline}~(b)}) to learn an auxiliary posterior distribution $q_{\phi}(z^{\prime} \vert X,C)$ subjecting to a Gaussian distribution, denoted as $\mathcal{N}(\mu_{z}^{\prime}, \sigma_z^{\prime2})$, to regularize $p_{\theta}(z \vert C)$, i.e., $p_{\theta}(z \vert C)$ should be close to $q_{\phi}(z^{\prime} \vert X,C)$.

Specifically, for the recognition network, we adopt the same learning architecture as the prior network to generate point cloud embeddings, which are concatenated with ground-truth bounding box information and fed into the subsequent MLP layers to learn $q_{\phi}(z^{\prime} \vert X,C)$. Moreover, to facilitate the learning process, we encode the information $X$ into offsets relative to predefined anchors, and then perform normalization as:
\begin{equation}\label{eq:box_param}
\begin{aligned}
%       t_{c_x} = \frac{c_x^{gt}}{d^a},~t_{c_y} = \frac{c_y^{gt}}{d^a}, ~t_{c_z} = \frac{c_z^{gt}}{h^a}, ~t_w = \log\frac{w^{gt}}{w^a},~t_l = \log\frac{l^{gt}}{l^a},~t_h = \log\frac{h^{gt}}{h^a}, t_r = \sin(r^{gt}),
%   \resizebox{0.92\textwidth}{!}{
~&t_{c_x} = \frac{c_x^{gt}}{d^a},~t_{c_y} = \frac{c_y^{gt}}{d^a},~t_{c_z} = \frac{c_z^{gt}}{h^a},\\
&t_w = \log\frac{w^{gt}}{w^a},~t_l = \log\frac{l^{gt}}{l^a},~t_h = \log\frac{h^{gt}}{h^a},\\
&t_r = \sin(r^{gt}),
%   }
\end{aligned}
% \vspace{-0.1cm}
\end{equation}
where ($w^a$,\,$l^a$,\,$h^a$) is the size of the predefined anchor located in the center of the point cloud, and $d^a=\sqrt{(l^a)^2+(w^a)^2}$ is the diagonal of the anchor box. We also take $\mathrm{cos}(r)$ as the additional input of the recognition network to handle the issue of angle periodicity.

\subsubsection{Objective Function}
%\JHdel{The GLENet adapted from CVAE is} 
Following CVAE~\citep{sohn2015learning}, we optimize GLENet by maximizing the variational lower bound of the conditional log-likelihood $p_\theta(X \vert C)$:
\begin{equation}\label{eq:ELBO}
\begin{aligned}
\log p_{\theta}(X \vert C)~\geq
& ~E_{q_{\phi}(z^{\prime} \vert X,C)}[\log p_{\theta} (X \vert z,C)]-\\
& KL(q_{\phi}(z^{\prime} \vert X,C) \vert\vert p_{\theta}(z \vert C)),
%\vspace{-0.1cm}
\end{aligned}
\end{equation}
where $E_q[p]$ returns the expectation of $p$ on the distribution of $q$, and $KL(\cdot)$ denotes KL-divergence. 

% Specifically, the first term $E_{q_{\phi}(z^{\prime} \vert X,C)}[\log p_{\theta}(X \vert z,C)]$ enforces latent variables to learn bounding box knowledge, while the second term $ KL(q_{\phi}(z^{\prime} \vert X,C) \vert p\vert _{\theta}(z \vert C))$ is aimed at regularizing the distribution of $z$ by minimizing the KL-divergence between $p_{\theta}(z \vert C)$ and $q_{\phi}(z^{\prime} \vert X,C)$.

%\JHdel{As formulated in Eq.~\eqref{eq:ELBO},} 
% The whole optimization objective of GLENet consists of a task term and a regularization term.
Specifically, the first term $E_{q_{\phi}(z^{\prime} \vert X,C)}[\log p_{\theta}(X \vert z,C)]$ enforces the prediction network to be able to restore ground-truth bounding box from latent variables. Following (\cite{yan2018second}) and~(\cite{deng2021voxel}), we explicitly define the bounding box reconstruction loss as
% \vspace{-0.3cm}
\begin{equation}
\label{GLENet_rec}
\begin{split}
L_{rec} = L_{rec}^{reg} + \lambda L_{rec}^{dir},
\end{split}
\end{equation}
where $L_{rec}^{reg}$ denotes the Huber loss imposed on the prediction and encoded regression targets as described in Eq.~\eqref{eq:box_param}, and $L_{rec}^{dir}$ denotes the binary cross-entropy loss used for direction classification.

The second term $ KL(q_{\phi}(z^{\prime} \vert X,C) \| p_{\theta}(z \vert C))$ is aimed at regularizing the distribution of $z$ by minimizing the KL-divergence between $p_{\theta}(z \vert C)$ and $q_{\phi}(z^{\prime} \vert X,C)$. Since $p_{\theta}(z \vert C)$ and $q_{\phi}(z^{\prime} \vert X,~C)$ are re-parameterized as $\mathcal{N}(\mu_{z}, \sigma_{z}^{2})$ and $\mathcal{N}(\mu_{z}^{\prime}$, $\sigma_z^{\prime2})$ through the prior network and the recognition network, respectively, we can explicitly define the regularization loss as:
% \vspace{-0.3cm}
\begin{equation}\label{GLENet_kl}
\resizebox{0.48\textwidth}{!}{
$L_{KL}(q_{\phi}(z^{\prime} \vert X,C) \|%\vert 
p_{\theta}(z \vert C)) = 
\mathrm{log}\dfrac{\sigma_z^{\prime}}{\sigma_z}
+ \dfrac{\sigma_z^2}{2\sigma_z^{\prime2}}
+ \dfrac{(\mu_{z}-\mu_{z}^{\prime})^2}{2\sigma_z^{\prime2}}$.
}
\end{equation}
%\begin{equation}\label{GLENet_kl}
%	\begin{split}
%		&D_{KL}(q_{\phi}(z \vert X,C) \vert p_{\theta}(z \vert C))\\ =&\mathrm{log}\dfrac{\sigma_z^{\prime}}{\sigma_z}
%		+ \dfrac{\sigma_z^2}{2\sigma_z^{\prime2}}
%		+ \dfrac{(\mu_{z}-\mu_{z}^{\prime})^2}{2\sigma_z^{\prime2}}.
%	\end{split}
%\end{equation}
Thus, the overall objective function is written as 
\begin{equation}
  L = L_{rec} + \gamma\,L_{KL},  
\end{equation}
where we empirically set the hyperparameter $\gamma$ to 1 in all experiments.

\section{Probabilistic 3D Detectors with Label Uncertainty} \label{bbr_label_uncertainty}

% \textcolor{blue}{In this part we will analyze the limitation of existing probabilistic 3D detectors and show how to introduce a unified way of integrating the label uncertainty estimated by GLENet into existing framework to build more powerful probabilistic detectors.}

%\subsection{Review of Probabilistic Object Detectors}

%Most existing probabilistic detectors model the prediction and ground-truth as Gaussian distribution and Dirac delta function, respectively
%Existing probabilistic detectors \cite{bbr} predict a probability distribution $P_{\Theta}(y)$ instead of only deterministic bounding box location:
%\JHNOTE{
% I think we can present this section in the following way: 
To reform a typical detector to be a probabilistic object detector, we can enforce the detection head to estimate a probability distribution over bounding boxes, denoted as $P_{\Theta}(y)$, instead of a deterministic bounding box location:%} %Instead of a deterministic bounding box location, existing probabilistic detectors \citep{bbr} estimate a probability distribution over bounding boxes $P_{\Theta}(y)$:
\begin{equation}
P_{\Theta}(y)=\frac{1}{\sqrt{2\pi\hat{\sigma}^2}}e^{-\frac{(y-\hat{y})^2}{2\hat{\sigma}^2}},
\end{equation}
where $\Theta$ indicates learnable network weights of a typical detector, $\hat{y}$ is the predicted bounding box location, and $\hat{\sigma}$ is the predicted localization variance.

Accordingly, we also assume the ground-truth bounding box as a Gaussian distribution $P_{D}(y)$ with variance $\sigma^2$, whose value is estimated by GLENet: %\textcolor{blue}{i.e., the label uncertainty estimated by GLENet:}
\begin{equation}
P_{D}(y)=\frac{1}{\sqrt{2\pi\sigma^2}}e^{-\frac{(y-y_g)^2}{2\sigma^2}},
\end{equation}
where $y_g$ represents the ground-truth bounding box.
%and approximate the \textcolor{blue}{label uncertainty $\sigma^2$} through GLENet. 
Therefore, we can incorporate the generated label uncertainty in the KL loss between the distribution of prediction and ground-truth in the detection head:
\begin{equation}\label{bbr_kl_loss}
\begin{aligned}
L_{reg} &= D_{KL}(P_{D}(y)||P_{\Theta}(y))\\
&= \mathrm{log}\dfrac{\hat{\sigma}}{\sigma} + \dfrac{\sigma^2}{2\hat{\sigma}^2}+\dfrac{(y_g-\hat{y})^2}{2\hat{\sigma}^2}.
\end{aligned}
\end{equation}

%\noindent where $y$ denotes the regression targets of detectors, $\hat{y}$ is the predicted offsets, and $\hat{\sigma}$ is the uncertainty of the estimation. Intuitively, given samples with high label uncertainty, the model is encouraged to predict larger variance ${\hat{\sigma}^2}$ under the supervision of $\sigma^2$.

\subsection{More Analysis of KL-Loss} \label{sec:theoretical analysis}
\begin{figure*}[htp]
\begin{subfigure}{.33\textwidth}
\centering
%		Contents of the sub-figure
\includegraphics[width=5cm]{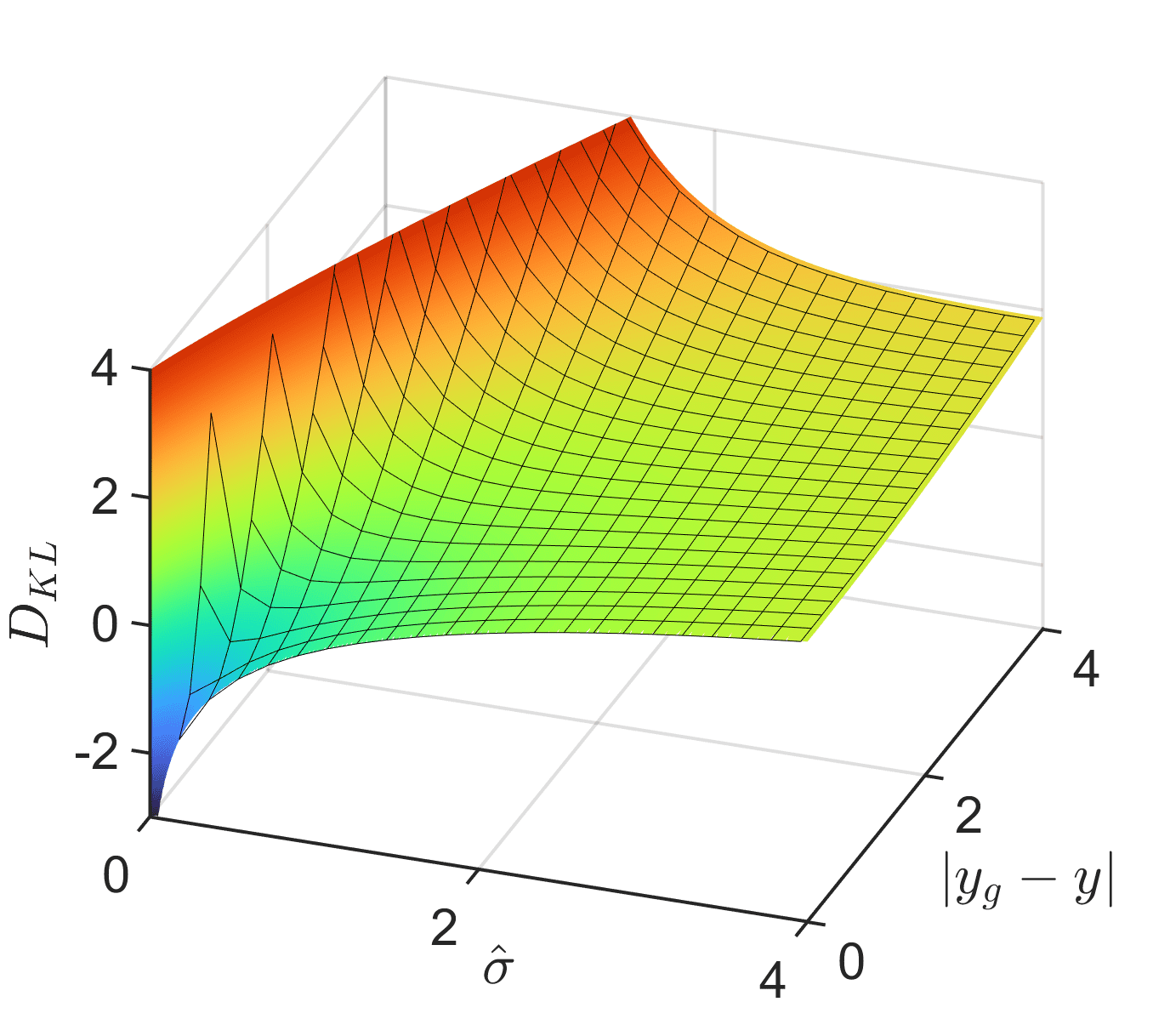}
\caption{$L_{reg}^{prob} (\sigma=0)$}
\label{kl_fig1}
\end{subfigure}
\begin{subfigure}{.33\textwidth}
\centering
%		Contents of the sub-figure
\includegraphics[width=5cm]{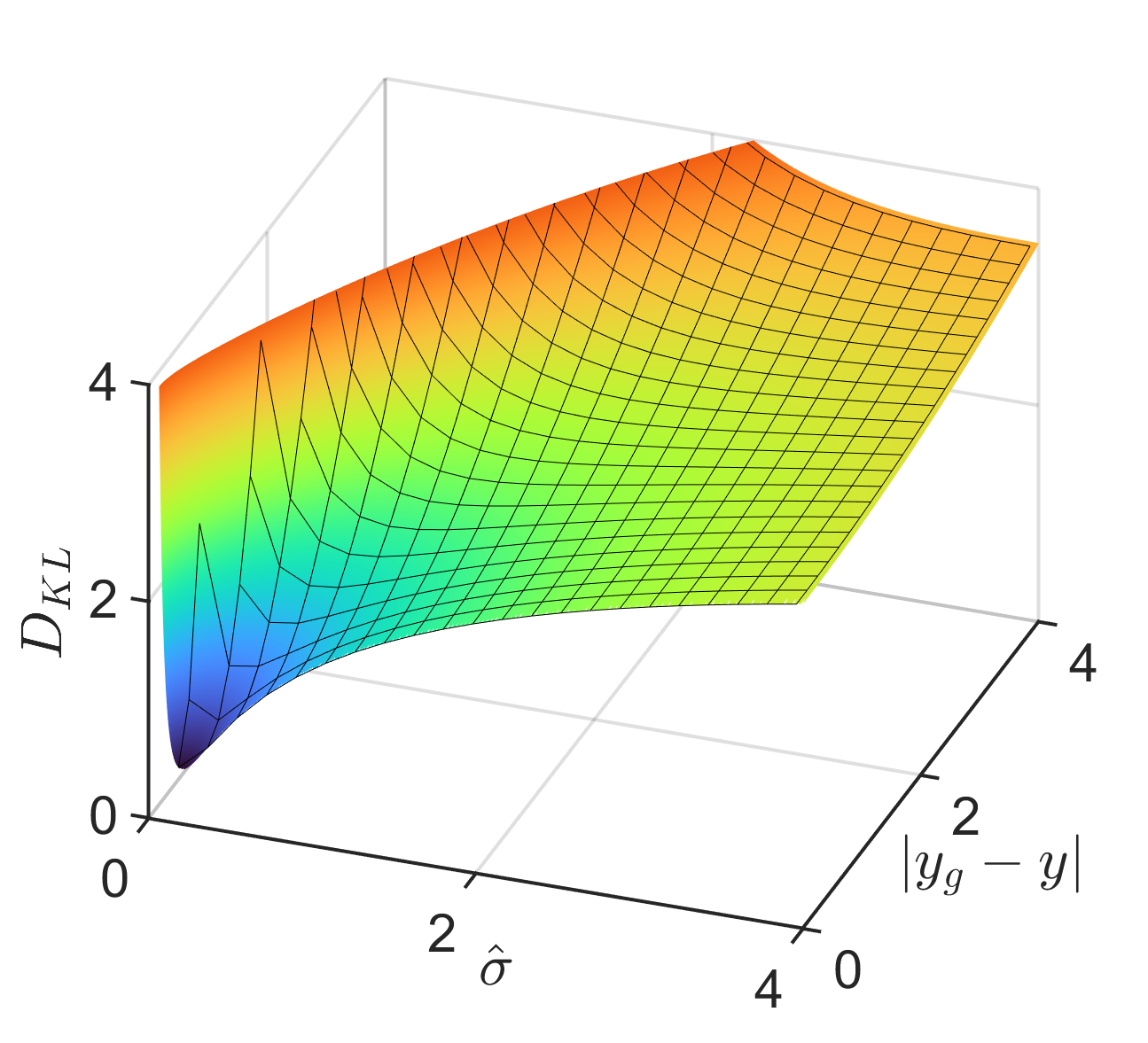}
\caption{$L_{reg} (\sigma=0.2)$}
\label{kl_fig2}
\end{subfigure}
\begin{subfigure}{.33\textwidth}
\centering
%		Contents of the sub-figure
\includegraphics[width=5cm]{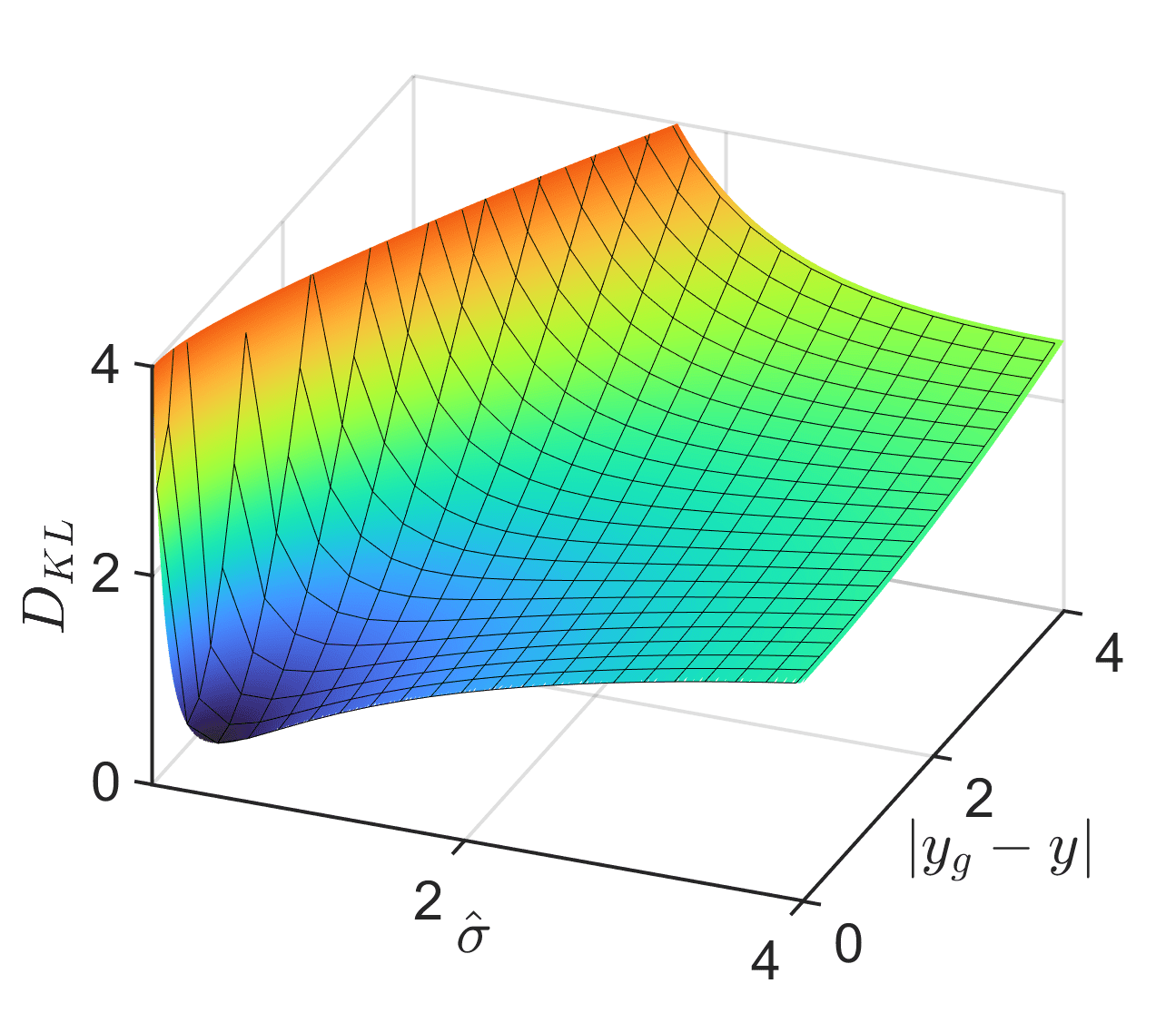}
\caption{$L_{reg} (\sigma=0.5)$}
\label{kl_fig3}
\end{subfigure}
\caption{Illustration of the KL-divergence between distributions as a function of localization error $|y_g-\hat{y}|$ and estimated localization variance $\hat{\sigma}$ given different label uncertainty $\sigma$. With label uncertainty $\sigma$ estimated by GLENet instead of zero, the gradient is smoother when the loss converges to the minimum. Besides, the $L_{reg}$ is smaller when $\sigma$ is larger, which prevents the model from overfitting to uncertain annotations.}
\label{fig:loss_surface}
\end{figure*}

When ignoring label ambiguity and formulating the ground-truth bounding box as a Dirac delta function, as done in (\cite{bbr}), the loss in Eq.~\eqref{bbr_kl_loss} degenerates into
\begin{equation}\label{old_kl_loss}
L_{reg}^{prob} 
% &= D_{KL}\left(P_{D}(y)||P_{\Theta}(y)\right)
\propto \dfrac{\mathrm{log}(\hat{\sigma}^2)}{2} + \dfrac{(y_g-\hat{y})^2}{2\hat{\sigma}^2},
\end{equation}
and the partial derivative of Eq.~\eqref{old_kl_loss} with respect to the predicted variance $\hat{\sigma}$ is:
\begin{equation}
\frac{\partial L_{reg}^{prob}}{\partial \hat{\sigma } } =\frac{1}{\hat{\sigma}} - \frac{(y_g-\hat{y})^2}{\hat{\sigma}^3}.
\end{equation}
When minimizing Eq.~\eqref{old_kl_loss}, a potential issue is that as $|y_g-\hat{y}|\to 0$,
\begin{equation}
\frac{\partial L_{reg}^{prob}}{\partial \hat{\sigma } } \to \frac{1}{\hat{\sigma}},
\end{equation}
% resulting in that the derivative for $\hat{\sigma}$ will explode when $\hat{\sigma}\to0$.
the derivative for $\hat{\sigma}$ can explode when $\hat{\sigma}\to0$.
Based on the property of KL-loss, the prediction is optimal only when the estimated $\hat{\sigma}=0$ and the localization error $|y_g-\hat{y}|=0$.
Therefore, the gradient explosion may result in erratic training and sub-optimal localization precision.

By contrast, after modeling the ground-truth bounding box as a Gaussian distribution, the partial derivative of Eq.~\eqref{bbr_kl_loss} with respect to prediction is:
\begin{equation}
\frac{\partial L_{reg}}{\partial \hat{\sigma } } =\frac{1}{\hat{\sigma}} -\frac{\sigma^2}{\hat{\sigma}^3} - \frac{(y_g-\hat{y})^2}{\hat{\sigma}^3},
\end{equation}
and
\begin{equation}
\frac{\partial L_{reg}}{\partial \hat{y} } = \frac{\hat{y}-y_g}{\hat{\sigma}^2}.
\end{equation}
As $|y_g-\hat{y}|\to0$ and $\hat{\sigma}>0$,
\begin{equation}
\frac{\revise{\partial L_{reg}}}{\partial \hat{\sigma } } \to \frac{1}{\hat{\sigma}}(1-\frac{\sigma^2}{\hat{\sigma}^2}),
\end{equation}
and
\begin{equation}
\frac{\revise{\partial L_{reg}}}{\partial \hat{y} } \to 0.
\end{equation}
Thus, when the predicted distribution reaches the optimal solution that is the distribution of ground-truth, i.e., $|y_g-\hat{y}|\to0$ and $\hat{\sigma}\to\sigma$, the derivatives for both $\hat{y}$ and $\hat{\sigma}$ become zero, which is an ideal property for the loss function and avoids the aforementioned gradient explosion issue.

Fig.~\ref{fig:loss_surface} shows the landscape of the KL-divergence loss function under different label uncertainty $\sigma$, which are markedly different in shape and property. 
The $L_{reg}^{prob}$ approaches infinitesimal and the gradient explodes as $|y_g-\hat{y}|\to0$ and $\hat{\sigma}\to0$. However, when we introduce the estimated label uncertainty and the predicted distribution is equal to the ground-truth distribution, the KL Loss has a determined minimum value of 0.5 and the gradient is smoother.

\begin{figure*}[htp]
\centering
\includegraphics[width=0.88\textwidth]{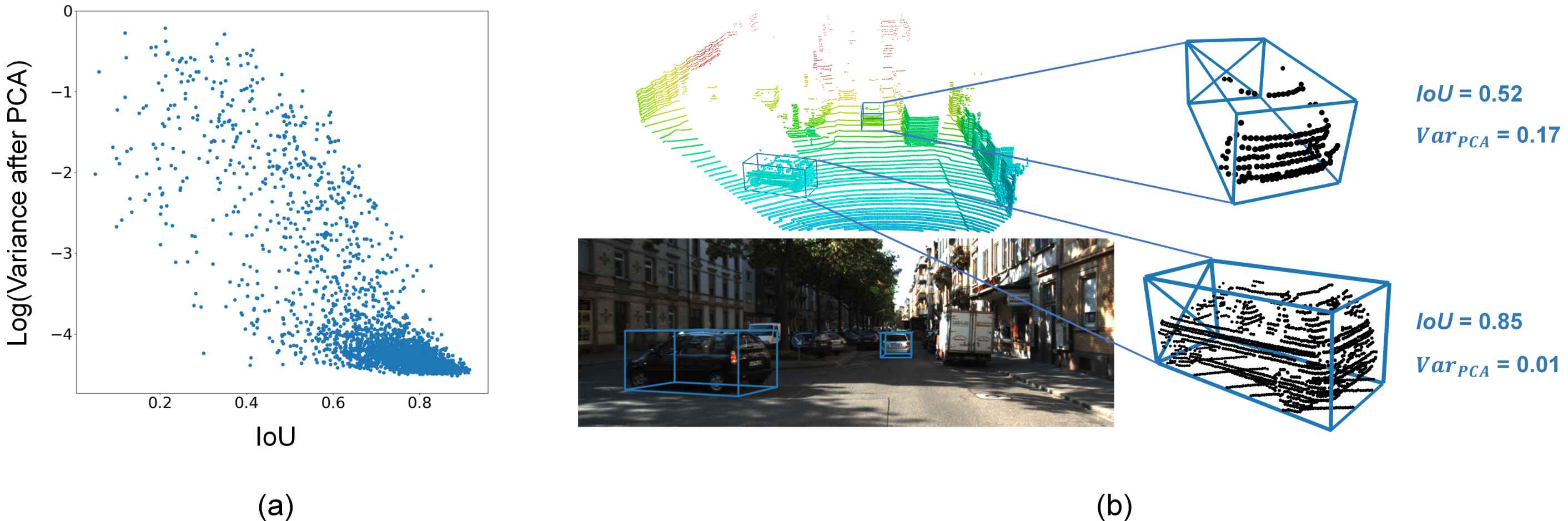} %\vspace{-0.3cm}
\caption{%\JHdel{Motivation of utilizing the learned uncertainty of bounding box distributions to facilitate the training of the IoU estimating branch.} 
(a) Illustration of the relationship between the actual localization precision (i.e., IoU between predicted and ground-truth bounding box) and the variance predicted by a probabilistic detector. Here, we reduce the dimension of the variance with PCA to facilitate visualization. (b) Two examples: for the sparse sample, the prediction has high uncertainty and low localization quality, while for the dense sample, the prediction has high localization quality and low uncertainty estimation.}
\label{fig:motivation_quality}
\end{figure*}

\subsection{Uncertainty-aware Quality Estimator}\label{sec:UAQE}

\begin{figure}[t]
\centering
\includegraphics[width=0.5\textwidth]{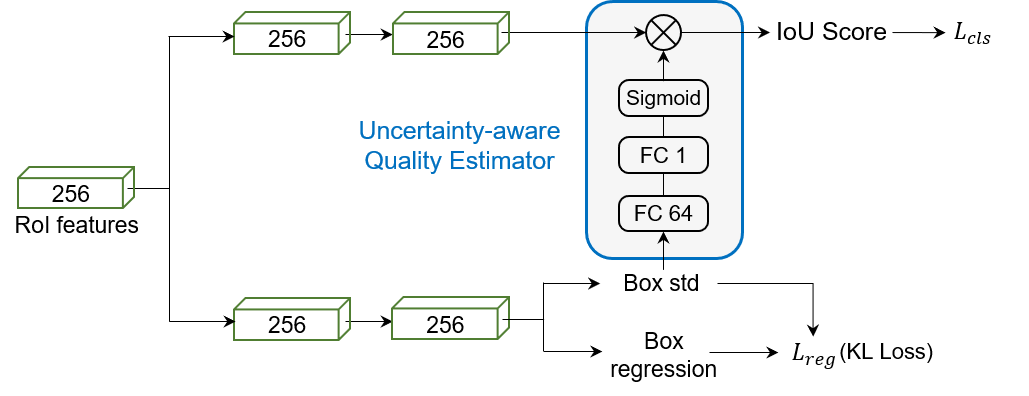}
\caption{Illustration of the proposed UAQE module in the detection head using the learned localization variance to assist the training of localization quality (IoU) estimation branch.}
\label{fig:head}
\end{figure}

Most state-of-the-art two-stage 3D object detectors use an IoU-related confidence score as the sorting criterion in NMS (non-maximum suppression), indicating the localization quality rather than the classification score.
As shown in Fig.~\ref{fig:motivation_quality}, there is a strong correlation between the uncertainty and actual localization quality for each bounding box. This observation motivates us to use uncertainty as a criterion for judging the quality of the boxes. However, the estimated uncertainty is 7-dimensional, making it infeasible to directly replace the IoU confidence score with the uncertainty.
To overcome this issue, we propose an uncertainty-aware quality estimator (UAQE) that introduces uncertainty information to facilitate the training of the IoU-branch and improve the accuracy of IoU estimation. The UAQE is shown in Fig. \ref{fig:head}. Given the predicted uncertainty as input, we construct a lightweight sub-module consisting of two fully connected (FC) layers followed by the Sigmoid activation to generate a coefficient. The original output of the IoU-branch is then multiplied with this coefficient to obtain the final estimation.
The UAQE aims to capture the uncertainty in the estimation and adjust the final output accordingly, resulting in a more accurate estimation of the IoU score. 

\subsection{3D Variance Voting}

Considering that in probabilistic object detectors, the learned localization variance by the KL loss can reflect the uncertainty of the predicted bounding boxes, following \cite{bbr}, we also propose 3D variance voting to combine neighboring bounding boxes to seek a more precise box representation. %representative. 
Specifically, at a single iteration in the loop, box $b$ with the maximum score is selected and its new location is calculated according to itself and the neighboring boxes.
During the merging process, the neighboring boxes that are closer and have a low variance are assigned higher weights. 
Note that neighboring boxes with a large angle difference from $b$ do not participate in the ensembling of angles. We refer the readers to Algorithm \ref{alg:3d_var_voting} for the details.
\begin{algorithm}[t]
%	\small
\SetAlgoLined
\caption{3D \revise{Variance Voting}}\label{alg:3d_var_voting}
\KwData{$B$ is an $N\times7$ matrix of predicted bounding boxes with parameters $(x,y,z,w,l,h,\theta)$. $C$ is the corresponding variance. $S$ is a set of N corresponding confidence values. $\sigma_t$ is a tunable hyperparameter.}
\KwResult{The final voting results $D$ of selected candidate boxes.}
$B=\{b_1,b_2,...,b_N\}$; and $C=\{c_1,c_2,...,c_N\}$\;
$S = \{s_1,s_2,...,s_N\}$; and $L=\{1,2,...,N\}$\;
$D\leftarrow$ \{\}\;
$iou_{thresh}\leftarrow \mu$\;
\While{$L\neq\emptyset$}{
idx =$\underset{i\in{L}}{\mathrm{argmax}}\,S, b'=b_{idx}$\;
$L'= \{i\vert i\in L, IoU(b_i, b') > iou_{thresh}\}$\;
$P\leftarrow$ \{\}\;
\For{$i\in L'$}{
$p_i = e^{-(1-IoU(b_i, b))^2/\sigma_t}$\;
\If{ $\vert tan(b_i^{\theta} - b^{'\theta})\vert >1$}{
	$p_i^{\theta}=0$\;
}
$P\leftarrow P\bigcup p_i$\;
}
$b_m = \frac{\sum_{i\in L'}{b_i\cdot p_i/c_i}}{\sum_{i\in L'}{p_i/c_i}}, p_i\in P, b_i\in B, c_i\in C$\;
$D\leftarrow D\bigcup b_m$\;
$L\leftarrow L - L'$\;	
}
\end{algorithm}

\vspace{-0.5cm}
\section{Experiments} \label{experiments}

To reveal the effectiveness and universality of our method, we integrated GLENet into several popular types of 3D object detection frameworks to form probabilistic detectors, which were evaluated on two commonly used benchmark datasets, i.e., the Waymo Open dataset (WOD) \citep{Sun_2020_CVPR} and the KITTI dataset \citep{Geiger_KITTI}. Specifically, we start by introducing specific experiment settings and implementation details in Section \ref{experiment_settings}. After that, we report the detection performance of the resulting probabilistic detectors and make comparisons with previous state-of-the-art approaches in Sections \ref{exp_on_kitti} and \ref{exp_on_waymo}. Finally, we conduct a series of ablation studies to verify the necessity of different key components and configurations in Section \ref{ablation_study}.

\begin{figure*}[htp]
\centering
\includegraphics[width=0.95\textwidth]{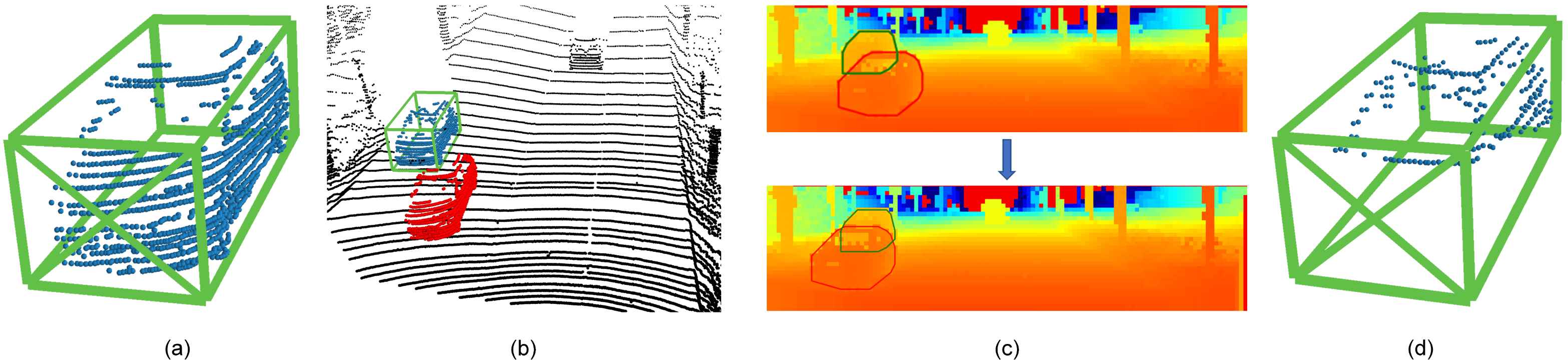}
\caption{%\JHNOTE{This figure and the caption should be revised.} 
Illustration of the occlusion data augmentation.
(a) The point cloud of the original object associated with the annotated ground-truth bounding box. (b) %Sample a dense object from the ground-truth database, and 
A sampled dense object (\textcolor{red}{red}) is placed between the LiDAR sensor and the original object (\textcolor{blue}{blue}). %\textcolor{blue}{The original and sampled objects are colored in blue and red, respectively.} 
(c) The projected range image from the point cloud in (b), %is projected to form a range image Project original and sampled objects to range image, 
where the convex hull (the red polygon) of the sampled object is calculated and further jittered to increase the diversity of occluded samples. Based on the convex hull (the \revise{green} polygon) of the original point cloud, the occluded area can be obtained.
The point cloud of the original object corresponding to the occluded area is removed. (d) Final augmented object with the annotated ground-truth bounding boxes.}
\label{fig:data_aug}
%	\vspace{-0.2cm}
\end{figure*}

\subsection{Experiment Settings} \label{experiment_settings}

\subsubsection{Benchmark Datasets} 
The \textbf{KITTI} dataset contains 7481 training samples with annotations in the camera field of vision and 7518 testing samples. According to the occlusion level, visibility, and bounding box size, the samples are further divided into three difficulty levels: simple, moderate, and hard. Following common practice, when performing experiments on the val set, we further split all training samples into a subset with 3712 samples for training and the remaining 3769 samples for validation. We report the performance on both the val set and online test leaderboard for comparison. And we use all training data for the test server submission.\\

\noindent The \textbf{Waymo Open} dataset is a large-scale autonomous driving dataset with more diverse scenes and object annotations in full $360^\circ$, which contains 798 sequences (158361 LiDAR frames) for training and 202 sequences (40077 LiDAR frames) for validation. These frames are further divided into two difficulty levels: LEVEL1 for boxes with more than five points and LEVEL2 for boxes with at least one point.
We report performance on both LEVEL 1 and LEVEL 2 difficulty objects using the recommended metrics, mean Average Precision (mAP) and mean Average Precision weighted by heading accuracy (mAPH). To conduct the experiments efficiently, we created a representative training set by randomly selecting 20\% of the frames from the original training set, which comprises approximately 32,000 frames. All evaluations were performed on the complete validation set, consisting of around 40,000 frames, using the official evaluation tool.
%\vspace{-0.4cm}

\subsubsection{Evaluation Metric for GLENet}
Due to the unavailability of the true distribution of a ground-truth bounding box, we propose to evaluate GLENet in a non-reference manner, in which the negative log-likelihood between the estimated distribution of ground-truth $p_{D}(X \vert C)$ subjecting to a Gaussian distribution $\mathcal{N}(\hat{t}, \sigma^2)$ and %prediction's distribution
$p_{\theta}(X \vert C)$ is computed:

\hspace{-0.7cm}
\resizebox{0.48\textwidth}{!}{
\begin{minipage}{\linewidth}
    \begin{align}\label{eval_metric}
    L_{NLL}(\theta) =& -\int p_{\theta}(X \vert C)\,\mathrm{log}\,p_{D}(X \vert C)\mathrm{d}X \\
    \approx& -\dfrac{1}{S}\sum_{i=1}^{S} \mathrm{log}\,p_{D}(X_i \vert C)\notag \\     
    =& -\dfrac{1}{S}\sum_{i=1}^{S} \sum_{k\in\{c_x,c_y,\atop c_z,w,l,h,r\}}\dfrac{(t_k^i-\hat{t}_k^{i})^2}{2{\sigma_k}^2} + \dfrac{\mathrm{log}(\sigma_k^2)}{2} + \dfrac{\mathrm{log}(2\pi)}{2},\notag
    \end{align}
\end{minipage}
}

\noindent where $S$ denotes the number of inference times, $X_i$ is the result of the $i$-th inference, and $\hat{t}_k^{i}$ and $t_k^i$ represent the regression targets and the predicted offsets, respectively. We estimate the integral by randomly sampling multiple prediction results via the Monte Carlo method.
Generally, the value of $L_{NLL}$ is small when GLENet outputs reasonable bounding boxes, i.e., predicting diverse plausible boxes with high variance for incomplete point cloud and consistent, precise boxes with low variance for high-quality point cloud, respectively.

\subsubsection{Implementation Details} 
\label{sec:imple details}
% We trained GLENet on all annotated objects in the training set.
To prevent data leakage, we kept the dataset division of GLENet consistent with that of the downstream detectors. 
As the initial input of GLENet, the point cloud of each object was uniformly pre-processed into 512 points via random subsampling/upsampling. Then we decentralized the point cloud by subtracting the coordinates of the center point to eliminate the local impact of translation.

Architecturally, we realized the prior network and recognition network with an identical PointNet structure consisting of three FC layers of output dimensions (64, 128, 512), followed by another FC layer to generate an 8-dim latent variable. To avoid posterior collapse, we particularly chose a lightweight PointNet structure with channel dimensions (8, 8, 8) in the context encoder. The prediction network concatenates the generated latent variable and context features and feeds them into subsequent FC layers of channels (64, 64) before predicting offsets and directions.

\subsubsection{Training and Inference Strategies} 
To optimize GLENet, we adopted Adam \citep{kingma2014adam} with a learning rate of 0.003, $\beta_1$ of 0.9, and $\beta_2$ of 0.99. The model was trained for a total of 400 epochs on KITTI and 40 epochs on Waymo, with a batch size of 64 on 2 GPUs. We used the one-cycle policy \citep{smith2017cyclical} to update the learning rate.

In the training process, we applied common data augmentation strategies, including random flipping, scaling, and rotation, in which the scaling factor and rotation angle were uniformly drawn from [0.95, 1.05] and $[-\pi/4, \pi/4]$, respectively. It is important to include multiple plausible ground-truth boxes in training, especially for incomplete point clouds, so we further propose an occlusion-driven augmentation approach, as illustrated in Fig.~\ref{fig:data_aug}, after which a complete point cloud may look similar to another incomplete point cloud, while the ground-truth boxes of them are completely different. To overcome posterior collapse, we also adopted KL annealing \citep{bowman2016generating} to gradually increase the weight of the KL loss from 0 to 1. We followed k-fold cross-sampling to divide all training objects into ten mutually exclusive subsets. To overcome overfitting, each time we trained GLENet on 9 subsets and then made predictions on the remaining subset to generate label uncertainty estimations on the whole training set. During inference, we sampled the latent variable $z$ from the predicted prior distribution $p_{\theta}(z \vert c)$ 30 times to form multiple predictions, the variance of which was used as the label uncertainty.

\setlength{\tabcolsep}{15pt}
\begin{table*}[t]
\centering
\caption{Quantitative comparison with state-of-the-art methods on the KITTI test set for vehicle detection, under the evaluation metric of 3D Average Precision (AP) of 40 sampling recall points. The best and second-best results are highlighted in bold and underlined, respectively.}
\label{table:kitti_test}
\begin{tabular}{c|c|c|cccc} 
\toprule
\multirow{2}{*}{Method} & \multirow{2}{*}{Reference} & \multirow{2}{*}{Modality} & \multicolumn{4}{c}{ 3D $\mathrm{AP}_{R40}$  }      \\
&                            &                           & Easy  & Mod. & Hard  & mAP    \\ 
\hline
%F-PointNet~\citep{qi2018frustum}                   & CVPR'18                   & RGB+LiDAR                 & 89.81 & 79.28    & 74.59 & 81.23  \\
%EPNet~\citep{huang2020epnet}                   & ECCV'20                   & RGB+LiDAR                 & 89.81 & 79.28    & 74.59 & 81.23  \\
%3D-CVF~\citep{yoo20203d}                  & ECCV'20                   & RGB+LiDAR                 & 89.2  & 80.05    & 73.11 & 80.79  \\ 
MV3D~\citep{chen2017multi}		               & CVPR'17                   & RGB+LiDAR                 & 74.97          & 63.63          & 54.00             & 64.20           \\
F-PointNet~\citep{qi2018frustum}                 & CVPR'18                   & RGB+LiDAR                 & 82.19          & 69.79          & 60.59          & 70.86           \\
MMF~\citep{liang2019multi}                     & CVPR'19                   & RGB+LiDAR                 & 88.40           & 77.43          & 70.22          & 78.68           \\
PointPainting~\citep{vora2020pointpainting}  & CVPR'20                  & RGB+LiDAR                 & 82.11          & 71.70           & 67.08          & 73.63           \\
CLOCs~\citep{pang2020clocs}                      & IROS'20                  & RGB+LiDAR                 & 88.94          & 80.67          & 77.15          & 82.25           \\
EPNet~\citep{huang2020epnet}    & ECCV'20                   & RGB+LiDAR                 & 89.81          & 79.28          & 74.59          & 81.23           \\
3D-CVF~\citep{yoo20203d}   & ECCV'20                   & RGB+LiDAR                 & 89.20           & 80.05          & 73.11          & 80.79           \\ 
\hline
STD~\citep{yang2019std}                     & ICCV'19                  & LiDAR                     & 87.95 & 79.71    & 75.09 & 80.92  \\
Part-A2~\citep{shi2020points}                 & TPAMI'20                 & LiDAR                     & 87.81 & 78.49    & 73.51 & 79.94  \\
3DSSD~\citep{yang20203dssd}                   & CVPR'20                  & LiDAR                     & 88.36 & 79.57    & 74.55 & 80.83  \\
SA-SSD~\citep{He_2020_CVPR}                  & CVPR'20                  & LiDAR                     & 88.80  & 79.52    & 72.30  & 80.21  \\
PV-RCNN~\citep{shi2020pv}                 & CVPR'20                  & LiDAR                     & 90.25 & 81.43    & 76.82 & 82.83  \\
PointGNN~\citep{shi2020point}       & CVPR' 20                     & LiDAR                     & 88.33    & 79.47   & 72.29 & 80.03 \\
Voxel-RCNN~\citep{deng2021voxel}     & AAAI'21                       & LiDAR                     & 90.90     & 81.62    & 77.06    & 83.19  \\
SE-SSD~\citep{zheng2021se}                  & CVPR'21                  & LiDAR                     & \underline{91.49} & \underline{82.54}    & 77.15 & \underline{83.73}  \\
VoTR~\citep{mao2021voxel}                    & ICCV'21                  & LiDAR                     & 89.90  & 82.09    & \textbf{79.14} & 83.71  \\
Pyramid-PV~\citep{Mao_2021_ICCV}              & ICCV'21                  & LiDAR                     & 88.39 & 82.08    & 77.49 & 82.65  \\
CT3D~\citep{sheng2021improving}              & ICCV'21                  & LiDAR                     & 87.83 & 81.77    & 77.16 & 82.25  \\ 
\hline
GLENet-VR (Ours)         & -     & LiDAR                     & \textbf{91.67} & \textbf{83.23}    & \underline{78.43} & \textbf{84.44}  \\
\bottomrule
\end{tabular}
% \vspace{-0.1cm}
\end{table*}
\setlength{\tabcolsep}{1.5pt}

\setlength{\tabcolsep}{12pt}
\begin{table*}[htp]
\centering
\caption{Quantitative comparison of different methods on the KITTI validation set for vehicle detection, under the evaluation metric of 3D Average Precision (AP) calculated with 11 sampling recall positions. The 3D APs under 40 recall sampling recall points are also reported for the moderate car class. The best and second-best results are highlighted in bold and underlined, respectively.}
\label{table:kitti_val}
%	\scalebox{0.85}{
\begin{tabular}{c|c|ccc|ccc} 
\hline
\multirow{2}{*}{Methods} & \multirow{2}{*}{Reference} & \multicolumn{3}{c|}{3D $\mathrm{AP}_{R11}$} & \multicolumn{3}{c}{3D $\mathrm{AP}_{R40}$}  \\
&                            & Easy  & Moderate & Hard                     & Easy  & Moderate & Hard                     \\ 
\hline
Part-$A^2$~\citep{shi2020points}~             & TPAMI'20                 & 89.47 & 79.47    & 78.54                    & -     & -        & -                      \\
3DSSD~\citep{yang20203dssd}~                  & CVPR'20                  & 89.71 & 79.45    & 78.67                    & -     & -        & -                      \\
SA-SSD~\citep{He_2020_CVPR}~                 & CVPR'20                  & 90.15 & 79.91    & 78.78                    & 92.23 & 84.30    & 81.36                  \\
PV-RCNN~\citep{shi2020pv}~                & CVPR'20                  & 89.35 & 83.69    & 78.70                    & 92.57 & 84.83    & 83.31                  \\
SE-SSD~\citep{zheng2021se}~                 & CVPR'21                  & \textbf{90.21} & 85.71    & \textbf{79.22}                    & 93.19 & \textbf{86.12}    & 83.31                  \\
VoTR~\citep{mao2021voxel}~                   & ICCV'21                  & 89.04 & 84.04    & 78.68                    & -     & -        & -                      \\
Pyramid-PV~\citep{Mao_2021_ICCV}~             & ICCV'21                  & 89.37 & 84.38    & 78.84                    & -     & -        & -                      \\
CT3D~\citep{sheng2021improving}~                   & ICCV'21                  & 89.54 & \underline{86.06}    & 78.99                    & 92.85 & 85.82    & \underline{83.46}                  \\ 
\hline
SECOND~\citep{yan2018second}~                 & Sensors'18               & 88.61 & 78.62    & 77.22                    & 91.16 & 81.99    & 78.82                  \\
GLENet-S~(Ours)         & -                          & 88.68 & 82.95    & 78.19                    & 91.73 & 84.11    & 81.35                  \\ 
\hline
CIA-SSD~\citep{zheng2021cia}~                & AAAI'21                  & \underline{90.04} & 79.81    & 78.80                    & \textbf{93.59} & 84.16    & 81.20                  \\
GLENet-C~(Ours)         & -                          & 89.82 & 84.59    & 78.78                    & 93.20 & 85.16    & 81.94                  \\ 
\hline
Voxel R-CNN~\citep{deng2021voxel}~            & AAAI'21                  & 89.41 & 84.52    & 78.93                    & 92.38 & 85.29    & 82.86                  \\
GLENet-VR~(Ours)        & -                          & 89.93 & \textbf{86.46}    & \underline{79.19}                    & \underline{93.51} & \underline{86.10}    & \textbf{83.60}                  \\
\hline
\end{tabular}
%	}
% \vspace{-0.1cm}
\end{table*}
\setlength{\tabcolsep}{1.5pt}

\subsubsection{Base Detectors} We integrated GLENet into \revise{four} popular deep 3D object detection frameworks, i.e., SECOND~\citep{yan2018second}, CIA-SSD~\citep{zheng2021cia}, \revise{CenterPoint (two-stage)}~\citep{yin2021center},  and Voxel R-CNN~\citep{deng2021voxel}, to construct probabilistic detectors, which are dubbed as GLENet-S, GLENet-C, \revise{GLENet-CP}, and GLENet-VR, respectively. Specifically, we introduced an extra FC layer on the top of the detection head to estimate standard deviations along with the box locations. Meanwhile, we applied the proposed UAQE to GLENet-VR to facilitate the training of the IoU-branch. \revise{Generally, we set the value of $\sigma_t$ to 0.05 and the value of $\mu$ to 0.01 in KITTI and 0.7 in Waymo dataset in 3D variance voting.} Note that for fair comparisons, we kept the network configurations of these base detectors unchanged except those related to the new submodules.

\subsection{Evaluation on the KITTI Dataset} \label{exp_on_kitti}
\begin{figure}[t]
	\centering
	\includegraphics[width=0.45\textwidth]{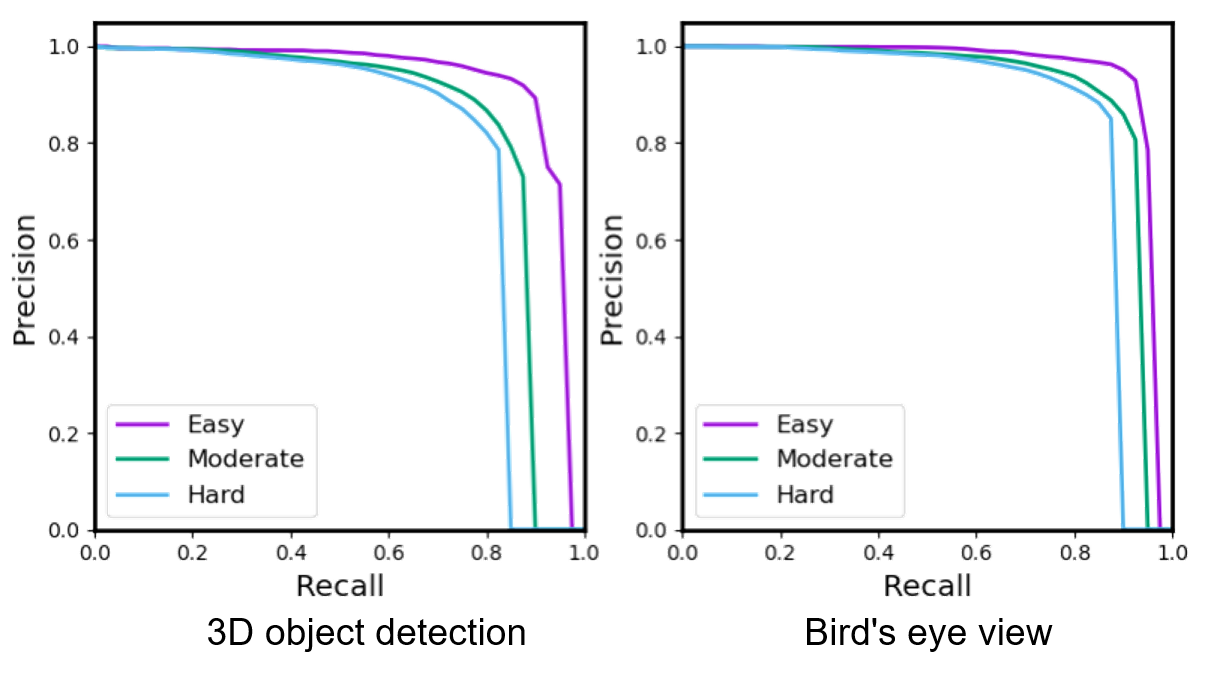}
	\caption{PR curves of GLENet-VR on the car class of the KITTI test set.
	}
	\label{fig:pr_curve}
\end{figure}

We compared GLENet-VR with state-of-the-art detectors on the KITTI test set, and Table~\ref{table:kitti_test} reports the AP and mAP that averages over the APs of easy, moderate and hard objects. As of March $29^{th}$, 2022, our GLENet-VR surpasses all published single-modal detection methods by a large margin and ranks \textbf{$1^{st}$} among all published LiDAR-based approaches. %\JHdel{on the highly competitive KITTI 3D detection benchmark}.  
Besides, Fig.~\ref{fig:pr_curve} also provides the detailed \revise{Precision}-Recall (PR) curves of GLENet-VR on KITTI test split.

\begin{table}[t]
	\centering
	\caption{\revise{Performance comparisons on the KITTI val set for pedestrian and cyclist class using $\mathrm{AP}_{R11}$.}}
	\label{table:kitti_val_3class}
	\begin{tabular}{>{\centering\hspace{0pt}}m{0.225\linewidth}|>{\centering\hspace{0pt}}m{0.092\linewidth}>{\centering\hspace{0pt}}m{0.156\linewidth}>{\centering\hspace{0pt}}m{0.092\linewidth}|>{\centering\hspace{0pt}}m{0.092\linewidth}>{\centering\hspace{0pt}}m{0.156\linewidth}>{\centering\arraybackslash\hspace{0pt}}m{0.092\linewidth}} 
		\toprule
		\multirow{2}{*}{\Centering{}Method} & \multicolumn{3}{>{\Centering\hspace{0pt}}m{0.339\linewidth}|}{Pedestrian} & \multicolumn{3}{>{\Centering\hspace{0pt}}m{0.339\linewidth}}{Cyclist}  \\
		& Easy  & Moderate & Hard                                                      & Easy  & Moderate & Hard                                                   \\ 
		\hline
		Second                              & 56.55 & \textbf{52.97}    & 47.73                                                     & 80.59 & 67.14    & 63.11                                                  \\
		GLENet-S                            & \textbf{58.22} & 52.39    & \textbf{49.53}                                                     & \textbf{82.67} & \textbf{68.29}    & \textbf{65.62}                                                  \\ 
		\hline
		Voxel R-CNN                         & \textbf{66.32} & 60.52    & 55.42                                                     & 86.62 & 70.69    & 66.05                                                  \\
		GLENet-VR                           & 66.18 & \textbf{62.05}    & \textbf{56.00}                                                     & \textbf{87.28} & \textbf{74.07}    & \textbf{70.90}                                                  \\
		\bottomrule
	\end{tabular}
\end{table}

\setlength{\tabcolsep}{6pt}
\begin{table*}[hbp]
        \begin{threeparttable}
	\centering
	\caption{Quantitative comparison of different methods on the Waymo validation set for vehicle detection. $\star$: experiment results re-produced with the code of OpenPCDet\tnote{a}. The best and second-best results are highlighted in bold and underlined, respectively.}
	% \vspace{-0.1cm}
	\label{table:waymo_val}
	%	\scalebox{0.81}{
		\begin{tabular}{c|ccccc|ccccc}
			\toprule
			\multirow{2}{*}{Methods} & \multicolumn{4}{c}{LEVEL\_1 3D mAP} & mAPH    & \multicolumn{4}{c}{LEVEL\_2 3D mAP} & mAPH     \\
			& Overall & 0-30m & 30-50m & 50m-inf  & Overall & Overall & 0-30m & 30-50m & 50m-inf  & Overall  \\ 
			\hline
			PointPillar~\citep{Lang_2019_CVPR}    & 56.62   & 81.01 & 51.75  & 27.94    & -       & -   & - & -  & -    & -        \\
			MVF~\citep{zhou2020end}    & 62.93   & 86.30 & 60.02  & 36.02    & -       & -       & -     & -      & -        & -        \\
			PV-RCNN~\citep{shi2020pv}  & 70.30   & 91.92 & 69.21  & 42.17    & 69.69   & 65.36   & 91.58 & 65.13  & 36.46    & 64.79    \\
			VoTr-TSD~\citep{mao2021voxel}  & 74.95   & 92.28 & 73.36  & 51.09    & 74.25   & 65.91   & -     & -      & -        & 65.29    \\
			Pyramid-PV~\citep{Mao_2021_ICCV}  & 76.30   & 92.67 & 74.91  & 54.54    & 75.68   & 67.23   & -     & -      & -        & 66.68    \\
			CT3D~\citep{sheng2021improving}   & 76.30   & 92.51 & 75.07  & 55.36    & -       & \underline{69.04}   & 91.76 & 68.93  & 42.60    & -        \\ 
			\hline
			SECOND${}^{\star}$~\citep{yan2018second}      & 69.85   & 90.71 & 68.93  & 41.17    & 69.40   & 62.76   & 86.92 & 62.57  & 35.89    & 62.30    \\
			GLENet-S~(Ours)          & 72.29   & 91.02 & 71.86  & 45.43    & 71.85   & 64.78   & 87.56 & 65.11  & 38.60    & 64.25    \\ 
			\hline
			CenterPoint-TS$\mathrm{}^\star$~\citep{yin2021center} & 75.52  & 92.09  & 74.35  & 54.27  & 75.07    & 67.37  & 90.89  & 68.11  & 42.46  & 66.94   \\
			GLENet-CP~(Ours)                       & \underline{76.73}  & \underline{92.70}  & \underline{75.70}  & \underline{55.77}  & \underline{76.27}    & 68.50  & \underline{91.95}  & 69.43  & \underline{43.68}  & \underline{68.08}   \\ 
			\hline
			Voxel R-CNN$\mathrm {}^{\star}$~\citep{deng2021voxel}  & 76.08   & 92.44 & 74.67  & 54.69    & 75.67   & 68.06   & 91.56 & \underline{69.62}  & 42.80    & 67.64    \\
			GLENet-VR~(Ours)         & \textbf{77.32}   & \textbf{92.97} & \textbf{76.28}  & \textbf{55.98}    & \textbf{76.85}   & \textbf{69.68}   & \textbf{92.09} & \textbf{71.21}  & \textbf{44.36}    & \textbf{68.97}    \\
			\bottomrule
		\end{tabular}
        \begin{tablenotes}
            \item[a] Reference: https://github.com/open-mmlab/OpenPCDet.
        \end{tablenotes}
        \end{threeparttable}
		%}
	% \vspace{-0.3cm}
\end{table*}
\setlength{\tabcolsep}{1.5pt}

Table~\ref{table:kitti_val} lists the validation results of different detection frameworks on the KITTI dataset, from which we can observe that GLENet-S, GLENet-C, and GLENet-VR consistently outperform their corresponding baseline methods, i.e., SECOND, CIA-SSD, and Voxel R-CNN, by 4.79\%, 4.78\%, and 1.84\% in terms of 3D R11 AP on the category of moderate car. Particularly, GLENet-VR achieves 86.36\% AP on the moderate car class, which surpasses all other state-of-the-art methods. Besides, as a single-stage method, GLENet-C achieves 84.59\% AP for the moderate vehicle class, which is comparable to the existing two-stage approaches while achieving relatively lower inference costs. It is worth noting that our method is compatible with mainstream detectors and can be expected to achieve better performance when combined with stronger base detectors.
\revise{Besides, our method also performs well on other classes. As shown in Table~\ref{table:kitti_val_3class}, GLENet-S outperforms the Second by +1.8\% and +2.51\% on pedestrian and cyclist classes respectively for 3D AP on the hard difficulty. And for the baseline Voxel R-CNN, our method improves the performance by +1.47\% and +3.38\% on pedestrian and cyclist classes respectively on the moderate difficulty.}
% \vspace{-0.3cm}

\subsection{Evaluation on the Waymo Open Dataset} \label{exp_on_waymo}
\revise{
In Table~\ref{table:waymo_val}, we present a comprehensive comparison of various state-of-the-art methods for vehicle detection on the Waymo Open Dataset, considering both LEVEL\_1 and LEVEL\_2 difficulty settings. 
The evaluation metrics used in this comparison include the 3D mean Average Precision (mAP) for different distance ranges (0-30m, 30-50m, and 50m-inf) and the overall mAP for LEVEL\_1 and LEVEL\_2.
Specifically, our method contributes 2.44\%, 1.21\% and 1.24\% improvements in terms of LEVEL\_1 mAP for SECOND, CenterPoint-TS and Voxel R-CNN, respectively. The improvements observed in the table demonstrate that our method is robust and consistently enhances the performance of baseline models like SECOND and Voxel R-CNN.
And GLENet-VR demonstrates the best performance with an mAP of 77.32\% and 69.68\% for LEVEL 1 and LEVEL 2, respectively. This superior performance can be attributed to the effective handling of bounding box ambiguity, especially for distant and sparse point cloud objects.
In addition to the overall performance, our methods also exhibit noteworthy improvements in the 30-50m and 50m-inf distance ranges. These results indicate that our method is particularly effective in resolving ambiguity for objects that are farther away from the sensor, which has traditionally posed challenges for point cloud-based detection algorithms.
In conclusion, Table~\ref{table:waymo_val} highlights the superior performance of our methods in 3D detection tasks on the Waymo Open Dataset. By effectively addressing the challenges posed by distant and sparse point cloud objects, our method demonstrates significant improvements in both LEVEL\_1 and LEVEL\_2 difficulty settings across various distance ranges.}

\subsection{Ablation Study} \label{ablation_study}
We conducted ablative analyses to verify the effectiveness and characteristics of our processing pipeline. In this section, all the involved model variants are built upon the Voxel R-CNN baseline and evaluated on the KITTI dataset, under the evaluation metric of average precision calculated with 40 recall positions.

\subsubsection{Comparison with Other Label Uncertainty Estimation} We compared GLENet with two other ways of label uncertainty estimation: 1) treating the label distribution as the deterministic Dirac delta distribution with zero uncertainty; 2) estimating the label uncertainty with simple heuristics, i.e., the number of points in the ground-truth bounding box or the IoU between the label bounding box and its convex hull of the aggregated LiDAR observations \citep{meyer2020learning}.
As shown in Table~\ref{table:label_generating}, our method consistently outperforms existing label uncertainty estimation paradigms. Compared with heuristic strategies, our deep generative learning paradigm can adaptively estimate label uncertainty statistics in 7 dimensions, instead of the uncertainty of bounding boxes as a whole, considering the variance in each dimension could be very different.
% [\textit{We should present more clearly that: 1) what ``the overall uncertainty of bounding boxes'' actually means? and 2) why it is not a good solution?}]}

%We did not 
Besides, to compare with \cite{feng_iros}, whose code is not publicly available, %To this end, 
we evaluated our method under its experiment settings and compared results with its reported performance.
As shown in Table~\ref{table:compare_35}, our method outperforms \cite{feng_iros} significantly in terms of $\mathrm{AP}_{BEV}$ on both moderate and hard levels.

\setlength{\tabcolsep}{5pt}
\begin{table}[t]
\centering
\caption{Comparison of different label uncertainty estimation approaches. "Convex hull" refers to the method in~\cite{meyer2020learning}. \revise{The best results are highlighted in bold.}}
% \vspace{-0.1cm}
\label{table:label_generating}
%\scalebox{0.8}{
\begin{tabular}{l|ccc} 
\toprule
% \multicolumn{1}{c|}{\multirow{2}{*}{Method}} & \multicolumn{3}{c}{3D $\mathrm{AP}_{R40}$}  \\
% \multicolumn{1}{c|}{}                         & Easy  & Moderate & Hard                     \\ 
\multirow{2}{*}{Methods}                    & \multicolumn{3}{c}{3D $\mathrm{AP}_{R40}$}  \\
                                           & Easy  & Moderate  & Hard                        \\ 
\hline
Voxel R-CNN                                   & 92.38 & 85.29    & 82.86                    \\
GLENet-VR w/~$L_{KLD}$~($\sigma^2$=0)           & 92.48 & 85.37    & 83.05                    \\
GLENet-VR w/~$L_{KLD}$~(points num)             & 92.46 & 85.58    & 83.16                    \\
GLENet-VR w/~$L_{KLD}$~(convex hull)   & 92.33 & 85.45    & 82.81                    \\
GLENet-VR w/~$L_{KLD}$~(Ours)                   & \textbf{93.49} & \textbf{86.10}    & \textbf{83.56}                    \\
\bottomrule
\end{tabular}
\end{table}
\setlength{\tabcolsep}{1.5pt}

\setlength{\tabcolsep}{4pt}
\begin{table}[htp]
\centering
\caption{Comparison of our method with ~\cite{feng_iros} on the KITTI val set. \revise{The best results are highlighted in bold.}}
\label{table:compare_35}
\begin{tabular}{l|ccc} 
\toprule
\multirow{2}{*}{Method}                    & \multicolumn{3}{c}{$AP_{BEV}$ for IoU@0.7}  \\
                                           & Easy  & Mod.  & Hard                        \\ 
\hline
PIXOR~\citep{yang2018pixor}                 					   & 86.79  & 80.75    & 76.60     \\
ProbPIXOR + $\mathcal{L}_{KLD}$ ($\sigma=0$)      & 88.60  & 80.44    & 78.74     \\
ProbPIXOR + $\mathcal{L}_{KLD}$ \citep{feng_iros}    & \textbf{92.22} & 82.03    & 79.16     \\
ProbPIXOR + $\mathcal{L}_{KLD}$ (Ours)          & 91.50  & \textbf{84.23}    & \textbf{81.85}     \\
\bottomrule
\end{tabular}
%	\vspace{-0.45cm}
\end{table}
\setlength{\tabcolsep}{1.5pt}

\subsubsection{Key Components of Probabilistic Detectors} We analyzed the contributions of different key components in our constructed probabilistic detectors and reported results in Table~\ref{table:components}. According to the second row, we can conclude that only training with the KL loss brings little performance gain. Introducing the label uncertainty generated by GLENet into the KL Loss contributes 0.75\%, 0.51\%, and 0.3\% improvements on the APs of easy, moderate, and hard classes, respectively, which demonstrates its regularization effect on KL-loss (Eq.~\ref{bbr_kl_loss}) and its ability to estimate more reliable uncertainty statistics of bounding box labels. The proposed UAQE module in the probabilistic detection head boosts the easy, moderate, and hard APs by 0.25\%, 0.19\% and 0.15\%, respectively, validating its effectiveness in estimating the localization quality.

\begin{figure}[htp]
	\centering
	\includegraphics[width=0.45\textwidth]{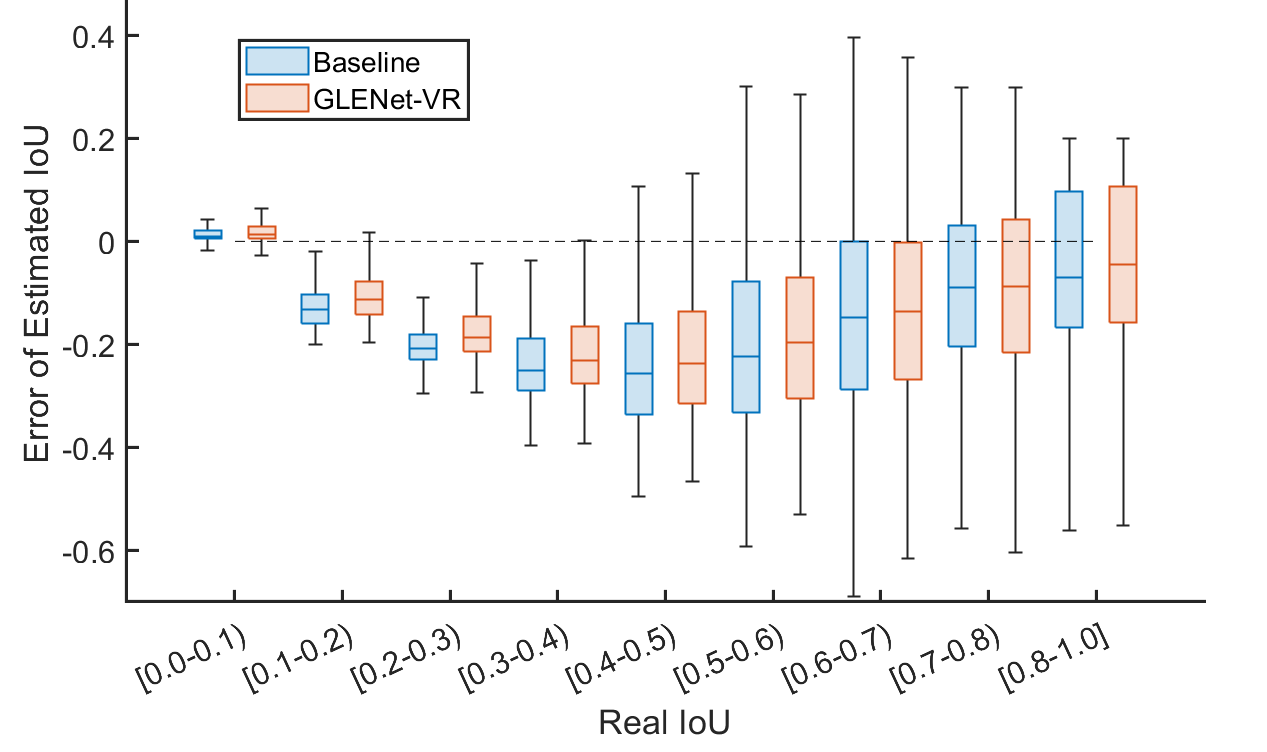}
	\caption{\revise{
 % Effect of UAQE for the IoU estimation of proposals. Specifically, we use Boxplot to show the distribution of error in the IoU estimation for proposals. 
Boxplots are used to display the estimated IoU error across various intervals of true IoU values. The x-axis represents the real IoU between proposals and their corresponding GT boxes, while the y-axis represents the distribution of estimation error, which is the difference between the estimated IoU score and the real IoU. The boxplot provides information about the distribution of error through five summary statistics: the minimum value, the maximum value, the median, the first quartile (Q1), and the third quartile (Q3).}}
	\label{fig:iou_dist_change}
\end{figure}

\revise{To gain a better understanding of how UAQE enhances the estimation of IoU-related confidence scores (the location quality), we analyze the error in IoU estimation for both GLENet-VR and the baseline model (w/o UAQE) over different actual IoU values between the proposals and their corresponding ground-truth boxes. Figure~\ref{fig:iou_dist_change} illustrates the changes in the error distribution of IoU estimation. We can observe that the UAQE module effectively reduces the IoU estimation error across various intervals of actual IoU values, such as [0.1, 0.6). These findings demonstrate that the UAQE module not only improves the overall average precision (AP) metric but also enhances the accuracy of location quality estimation.
}

\setlength{\tabcolsep}{5pt}
\begin{table}[t]
\centering
\caption{Contribution of each component in our constructed GLENet-VR pipeline. ``LU" denotes the label uncertainty.}
\label{table:components}
%	\vspace{-0.3cm}
%\scalebox{0.85}{
\begin{tabular}{cccc|ccc} 
\toprule
KL loss & LU & var voting & UAQE & Easy  & Moderate & Hard   \\ 
\hline
&    &        &      & 92.38 & 85.29    & 82.86  \\
\checkmark       &    &        &      & 92.45 & 85.25    & 82.99  \\
\checkmark       &    & \checkmark      &      & 92.48 & 85.37    & 83.05  \\
\checkmark       & \checkmark  &        &      & 93.20 & 85.76    & 83.29  \\
\checkmark       & \checkmark  & \checkmark      &      & 93.24 & 85.91    & 83.41  \\
\checkmark       & \checkmark  & \checkmark      & \checkmark    & \textbf{93.49} & \textbf{86.10}    & \textbf{83.56}  \\
\bottomrule
\end{tabular}
\end{table}
\setlength{\tabcolsep}{1.5pt}

\subsubsection{Ablation Study of GLENet}

\noindent
\revise{
	\textbf{Effectiveness of Preprocessing.} As mentioned previously, to eliminate the local impact of translation on the input of GLENet, the point cloud of a single object is standardized to zero mean value. However, this process might remove meaningful information contained in distances. For instance, distant objects with fewer points typically have high label uncertainty, while closer objects usually have a high point count and low label uncertainty.
	For this reason, we performed experiments by adding the absolute coordinates of the point cloud as an extra feature in the input of GLENet. However, as shown in Table~\ref{table:ablation_input}, the inclusion of extra absolute coordinates did not yield any significant improvement in the $L_{NLL}$ metric or the performance of downstream detectors.
	We reason these observations from two aspects. First, the additional absolute coordinates may differentiate objects that are located in different positions but have similar appearances. As a result, there may be fewer samples with similar shapes but different bounding box labels, making it difficult for GLENet to capture the one-to-many relationship between incomplete point cloud objects and potential plausible bounding boxes. Second, the absolute distance and the point cloud density are generally correlated, i.e., an object with a larger absolute distance generally has a sparser point cloud representation, and such correlation could be perceived by the network. In other words, the absolute distance information is somewhat redundant to the network.
}\\

\setlength{\tabcolsep}{5pt}
\begin{table}[t]
	\centering
	\caption{\revise{Effect of point cloud input with and without absolute coordinates in GLENet. ``NC" denotes normalized coordinates of the partial point cloud, and ``AC" denotes absolute coordinates. We report the $L_{NLL}$ for evaluation of GLENet and the 3D average precisions of 40 sampling recall points for evaluation of downstream detectors.}}
	\label{table:ablation_input}
	\begin{tabular}{cc|c|cccc} 
		\toprule
		NC & AC &  $L_{NLL}$$\downarrow$       & Easy  & Mod.  & Hard  & Avg    \\ 
		\hline
		\checkmark                    &                        & \textbf{91.50}    & \textbf{93.49} & \textbf{86.10} & \textbf{83.56} & \textbf{87.72}  \\
		\checkmark                    & \checkmark                      & 147.33 & 93.21 & 85.66 & 83.35 & 87.41  \\
		\bottomrule
	\end{tabular}
\end{table}
\setlength{\tabcolsep}{1.5pt}

\noindent\textbf{Influence of Data Augmentation.} To generate similar point cloud shapes with diverse ground-truth bounding boxes during training of GLENet, we proposed an occlusion data augmentation strategy and generated more incomplete point clouds while keeping the bounding boxes unchanged (see Fig. \ref{fig:data_aug}).
As listed in Table~\ref{table:data_aug}, it can be seen that the occlusion data augmentation effectively enhances the performance of GLENet and the downstream detection task.\\ 
% Besides, the effectiveness of the $L_{NLL}$ metric is also validated, which is proposed to evaluate GLENet and select optimal configurations to generate reliable label uncertainty.

\setlength{\tabcolsep}{2.8pt}
\begin{table}
	\centering
	\caption{Ablation study on occlusion augmentation techniques \revise{and context encoder} in GLENet, in which we report the $L_{NLL}$ for evaluation of GLENet and the 3D average precisions of 40 sampling recall points for evaluation of downstream detectors.}
	\label{table:data_aug}
	\begin{tabular}{c|c|cccc} 
		\toprule
		Setting                    & $L_{NLL}$$\downarrow$ & Easy  & Mod.  & Hard  & Avg.   \\ 
		\hline
		Baseline                   & \textbf{91.50}  & \textbf{93.49} & \textbf{86.10} & \textbf{83.56} & \textbf{87.72}  \\
		w/o Occlusion Augmentation & 230.10 & 92.96 & 85.52 & 83.07 & 87.18  \\
		\revise{w/o Context Encoder}        & 434.93 & 92.65 & 85.31 & 82.59 & 86.85  \\
		\bottomrule
	\end{tabular}
\end{table}

\setlength{\tabcolsep}{1.5pt}

\noindent
\revise{
	\textbf{Necessity of the Context Encoder.} 
%	From a certain point of view, in addition to learning the distribution of latent variables, the prior and recognition network are also able to extract features of point cloud. In order to verify the necessity of the context encoder, we performed an ablation experiment for it.	As shown in Table~\ref{table:data_aug}, after removing context encoder, we observe that both $L_{NLL}$ metric and AP of downstream detector deteriorate significantly. The results demonstrate the necessity of context encoder to extract geometric features from point cloud and free the recognition and prior network to capture the underlying structure of input data in a low-dimensional space.
	In addition to learning the distribution of latent variables, the prior and recognition networks are also capable of extracting features from point clouds.
	To verify the necessity of the context encoder that is responsible for encoding contextual information from the input data in GLENet, we conducted an ablation experiment. 
%	The context encoder is responsible for encoding contextual information into the input data to improve the quality of the reconstructed output.
	As shown in Table~\ref{table:data_aug}, after removing the context encoder, we observed a significant deterioration in both the $L_{NLL}$ metric and the average precision (AP) of the downstream detector. 
	These results clearly demonstrate the necessity of the context encoder to extract geometric features from point clouds and allow the recognition and prior networks to focus on capturing the underlying structure of the input data in a low-dimensional space. Without the context encoder, the recognition and prior networks would need to learn both the geometric features and the contextual information from the input data, which would lead to poorer performance.
}\\

\setlength{\tabcolsep}{7pt}
\begin{table}
	\centering
	\caption{\revise{Ablation study of the dimensions of latent variables in GLENet.}}
	\label{table:ablation_dim}
	\begin{tabular}{c|c|cccc} 
		\toprule
		Dimensions & $L_{NLL}$$\downarrow$ & Easy  & Mod.  & Hard  & Avg.   \\ 
		\hline
		2          & 856.48 & 92.05 & 84.69 & 82.22 & 86.32  \\
		4          & 605.11 & 92.25 & 85.11 & 82.24 & 86.53  \\
		8          & 91.50  & \textbf{93.49} & \textbf{86.15} & 83.56 & \textbf{87.73}  \\
		32         & \textbf{86.16}  & 93.28 & 85.94 & \textbf{83.60} & 87.60  \\
		64         & 110.49 & 93.11 & 85.51 & 83.27 & 87.30  \\
		128        & 105.93 & 92.74 & 85.82 & 83.10 & 87.22  \\
		\bottomrule
	\end{tabular}
\end{table}
\setlength{\tabcolsep}{1.5pt}

\noindent
\revise{
	\textbf{Dimension of the Latent Variable.} Table.~\ref{table:ablation_dim} shows the performance of adopting latent variables with various dimensions for GLENet. We can observe that the accuracy increase gradually, with the dimensions of latent variables from 2 to 8, and the setting of 32-dimensional latent variables achieve similar performance. The results demonstrate a too-small dimension of the latent variables makes the GLENet unable to fully represent the underlying structure of the input data. 
    And setting the dimension of latent variables to larger values like 64 or 128 can lead to over-fitting and slight decreases in performance. When the dimension of the latent variables is too large, the model can easily memorize the noise and details in the training data, which is not helpful for generating new and useful samples.
    Besides, though the setting of 32-dim latent variables leads to the lowest $L_{NLL}$, the performance of downstream detectors is best using label uncertainty with 8-dim latent variables. Therefore, though the $L_{NLL}$ metric can reflect the quality of generating of GLENet to some extent, it is not guaranteed to be strongly correlated with the performance of downstream detectors.
}\\

\noindent
\revise{
	\textbf{Effects of the Sampling Times.} In Table~\ref{table:ablation_sampling_times}, we investigate the effects of the sampling times to calculate label uncertainty. We can observe that larger sampling times generally achieve lower $L_{NLL}$ and better performance of downstream detectors, and similar performance is observed when using more than 30 sampling times. Statistically speaking, the variance obtained after a certain number of sampling times will tend to stabilize.
	Hence, to balance the computation cost and performance, we empirically choose to calculate the label uncertainty with predicted multiple bounding boxes by sampling the latent variables 30 times.
}

\setlength{\tabcolsep}{7pt}
\begin{table}
	\centering
	\caption{\revise{Ablation study of the sampling times to calculate label uncertainty in GLENet.}}
	\label{table:ablation_sampling_times}
	\begin{tabular}{c|c|cccc} 
		\toprule
		Times & $L_{NLL}$$\downarrow$ & Easy  & Mod.  & Hard  & Avg.   \\ 
		\hline
		4   & 608.82 & 92.54 & 85.11 & 81.21 & 86.29  \\
		8   & 240.08 & 92.96 & 85.52 & 82.80 & 87.09  \\
		16  & 148.21 & 92.99 & 85.66 & 83.35 & 87.33  \\
		30  & 91.5   & 93.49 & 86.10 & \textbf{83.56} & \textbf{87.72}  \\
		64  & 86.76  & 93.37 & \textbf{86.16} & 83.42 & 87.65  \\
		128 & \textbf{77.06}  & \textbf{93.53} & 85.92 & 83.47 & 87.64  \\
		\bottomrule
	\end{tabular}
\end{table}
\setlength{\tabcolsep}{1.5pt}

\subsubsection{Conditional Analysis}\label{con_analysis}
To figure out in what cases our method improves the base detector most, we evaluated GLENet-VR on different occlusion levels and distance ranges. %, and compared it with the base detector. 
As shown in Table~\ref{table:cond_analysis}, compared with the baseline, our method mainly improves on the heavily occluded and distant samples, which suffer from more serious boundary ambiguities of ground-truth bounding boxes. 

\setlength{\tabcolsep}{2pt}
\begin{table}[t]
	\begin{minipage}{\linewidth}
%		\scalebox{0.75}{
			\begin{threeparttable}
			\caption{Comparison on different occlusion levels and distance ranges\tnote{a}, evaluated by the 3D Average Precision (AP) calculated with 40 sampling recall positions on the KITTI val set.}
			\centering
			\label{table:cond_analysis}
				\begin{tabular}{cc|cc|c} 
					\toprule
%					\multicolumn{2}{c|}{Method} & Voxel R-CNN~\citep{deng2021voxel} & GLENet-VR (Ours)   & Improvement    \\ 
					\multicolumn{2}{c|}{\thead{Methods}} & \thead{Voxel R-CNN \\ ~\citep{deng2021voxel}} & \thead{GLENet-VR \\ (Ours)}   & \thead{Improvement}    \\
					\hline
					\multirow{3}{*}{Occlusion\tnote{b}} & 0      & 92.35 & 93.51 & +\textsl{1.16}  \\
					& 1  & 76.91      & 78.64 & +\textsl{1.73}  \\
					& 2  & 54.32      & 56.93 & +\textsl{2.61}  \\ 
					\hline
					\multirow{3}{*}{Distance} & 0-20m  & 96.42      & 96.69 & +\textsl{0.27}  \\
					& 20-40m  & 83.82      & 86.87 & +\textsl{3.05}  \\
					& 40m-Inf  & 38.86      & 39.82 & +\textsl{0.96}  \\
					\bottomrule
				\end{tabular}

		     	\begin{tablenotes}
					\item[a] The results include separate APs for objects belonging to different occlusion levels and APs for the moderate vehicle class in different distance ranges.
					\item[b] Definition of occlusion levels: levels 0, 1 and 2 correspond to fully visible samples, partly occluded samples, and samples difficult to see respectively.
				\end{tablenotes}
				
			\end{threeparttable}
%		}
	\end{minipage}
\end{table}
\setlength{\tabcolsep}{1.4pt}

\begin{figure*}[t]
\centering
\includegraphics[width=\textwidth]{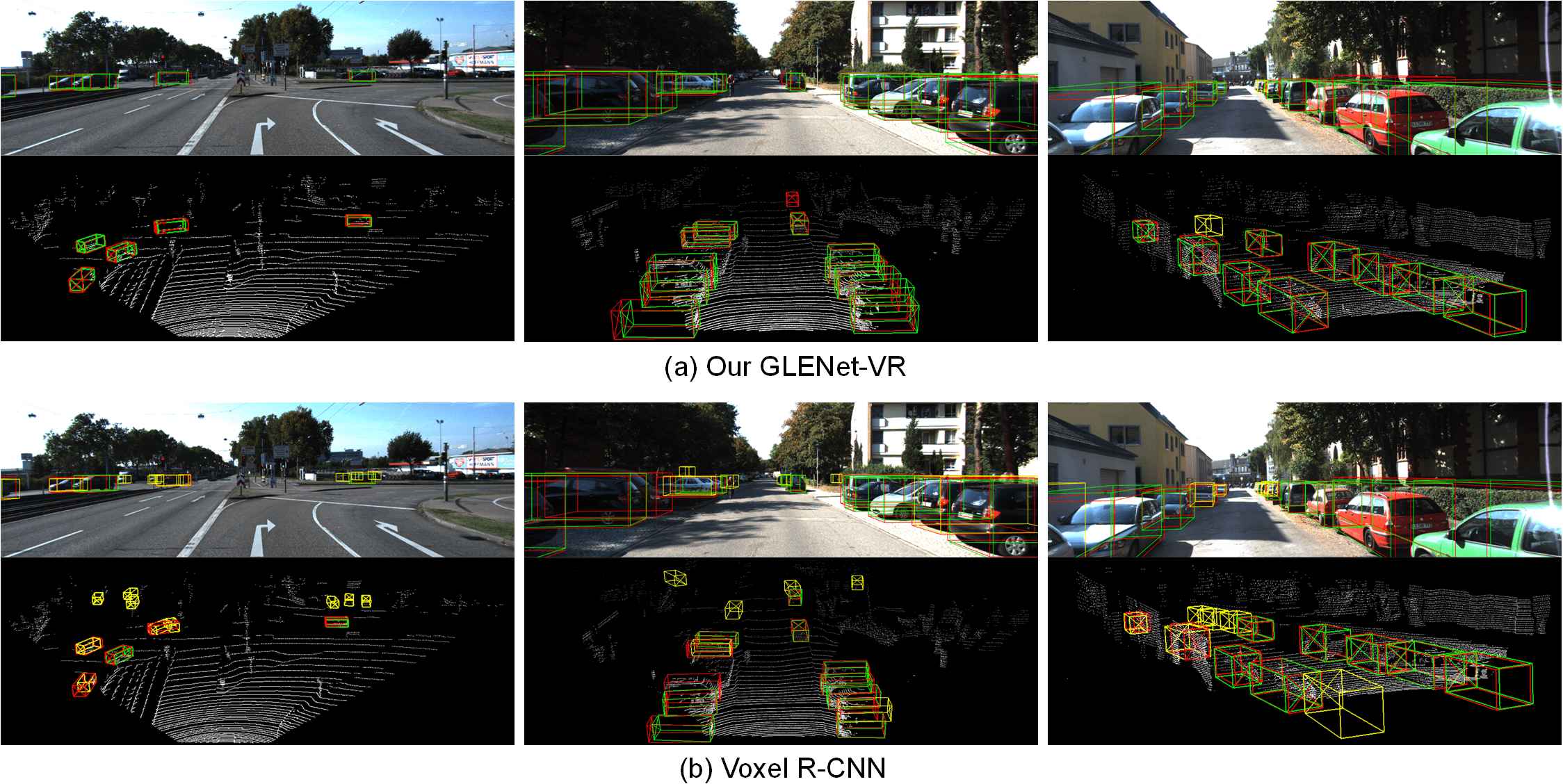}
\caption{Visual comparison of the results by GLENet-VR and Voxel R-CNN on the KITTI dataset. The ground-truth, true positive and false positive bounding boxes are visualized in red, green and yellow, respectively, on both the point cloud and image. Best viewed in color.
%		We also project the 3D bounding boxes back to the color images for visualization.
}
\label{fig:sm_debug_compare}
\end{figure*}

\begin{figure*}[htp]
	\centering
	\begin{minipage}{0.49\linewidth}
		\vspace{1pt}  
		\centerline{\includegraphics[width=\textwidth, height=4.4cm]{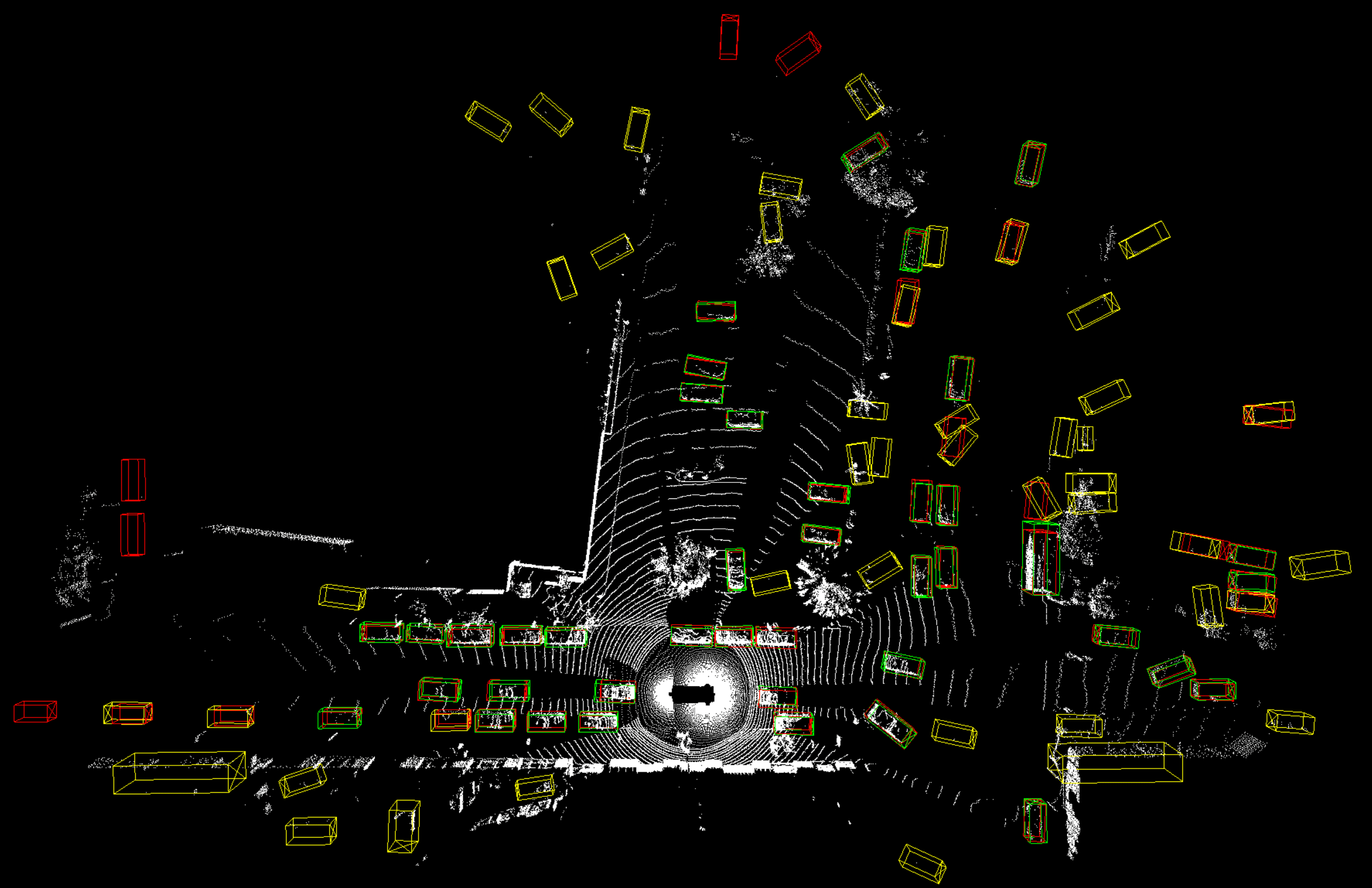}}
		\vspace{1pt}
		\centerline{\includegraphics[width=\textwidth, height=4.4cm]{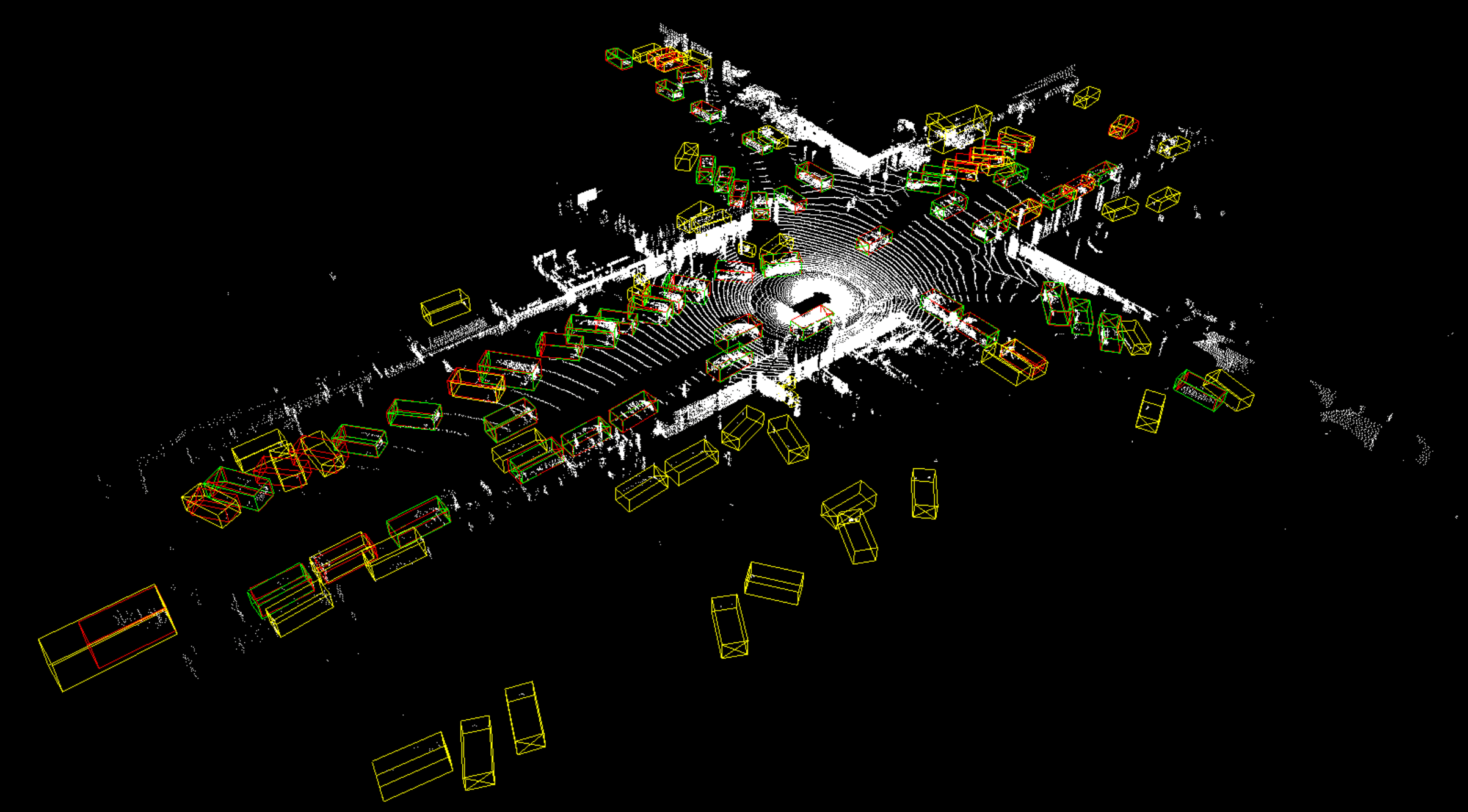}}
		\vspace{1pt}
		\centerline{\includegraphics[width=\textwidth, height=4.4cm]{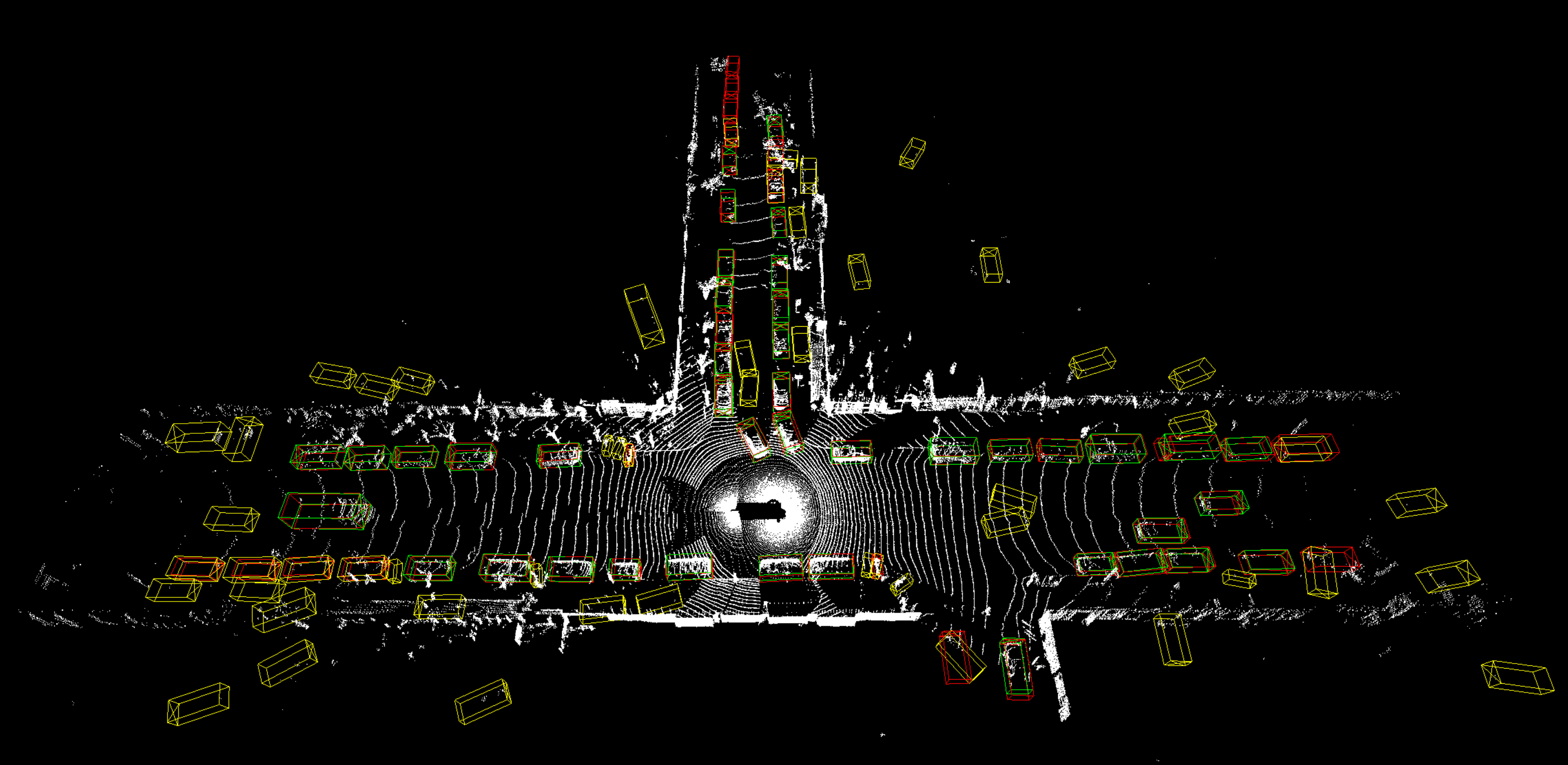}}
		\vspace{1pt}
		\centerline{(a) SECOND}
	\end{minipage}
%	\hspace{.01in}
	%一个 \begin{minipage}{0.49\linewidth} 就是一列
	\begin{minipage}{0.49\linewidth}
		\vspace{1pt}
		\centerline{\includegraphics[width=\textwidth, height=4.4cm]{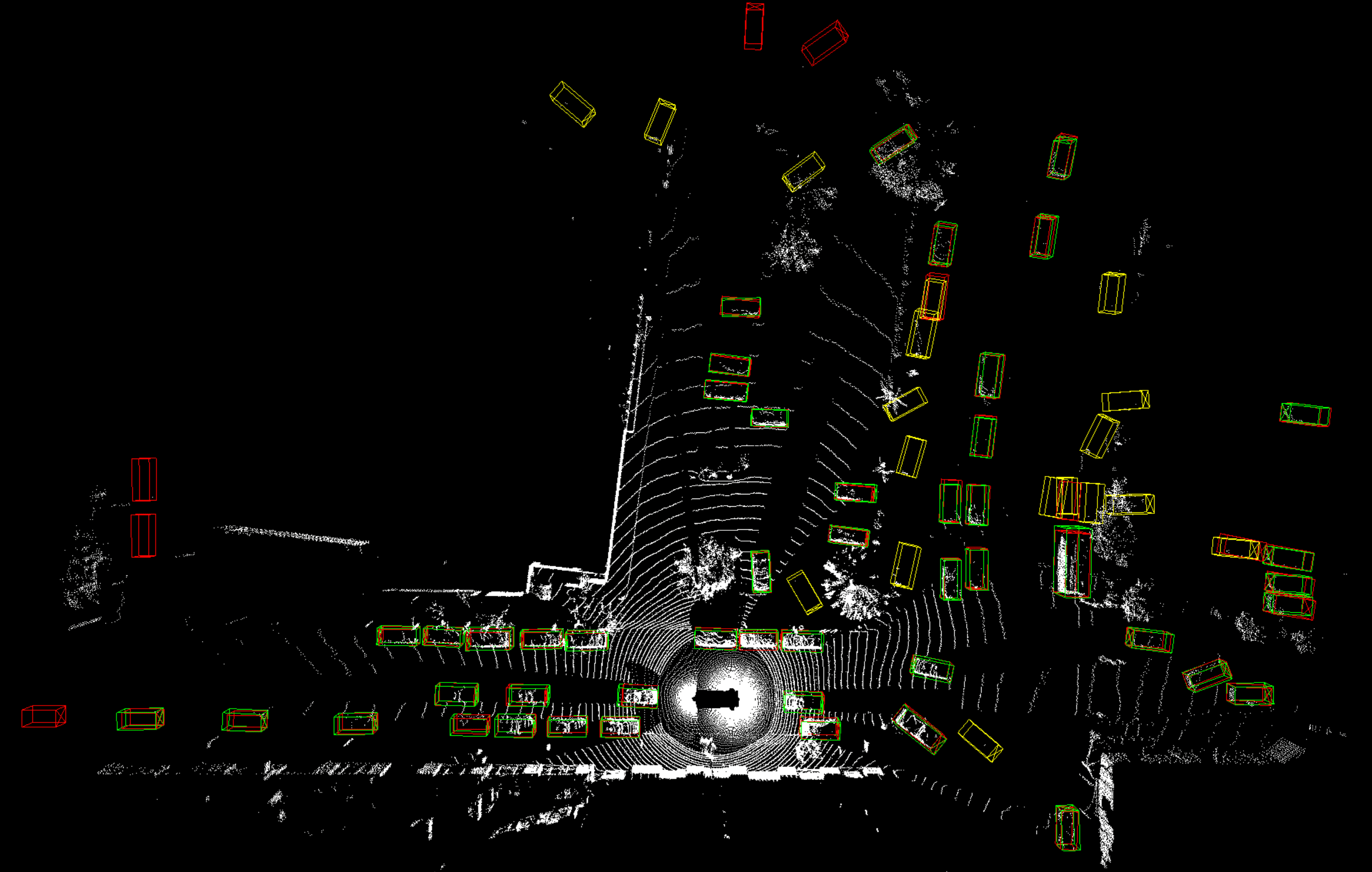}}
		\vspace{1pt}
		\centerline{\includegraphics[width=\textwidth, height=4.4cm]{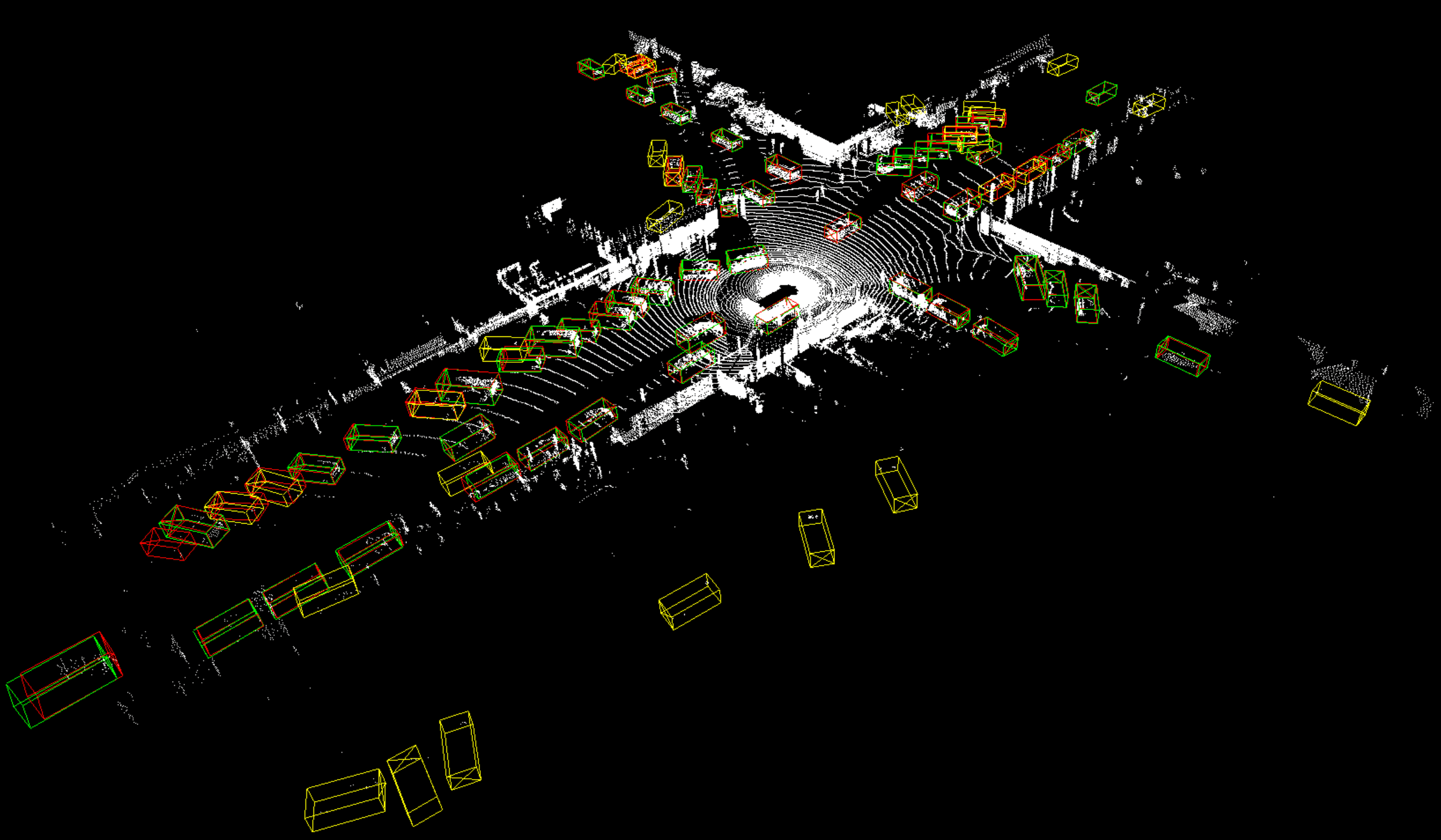}}
		\vspace{1pt}
		\centerline{\includegraphics[width=\textwidth, height=4.4cm]{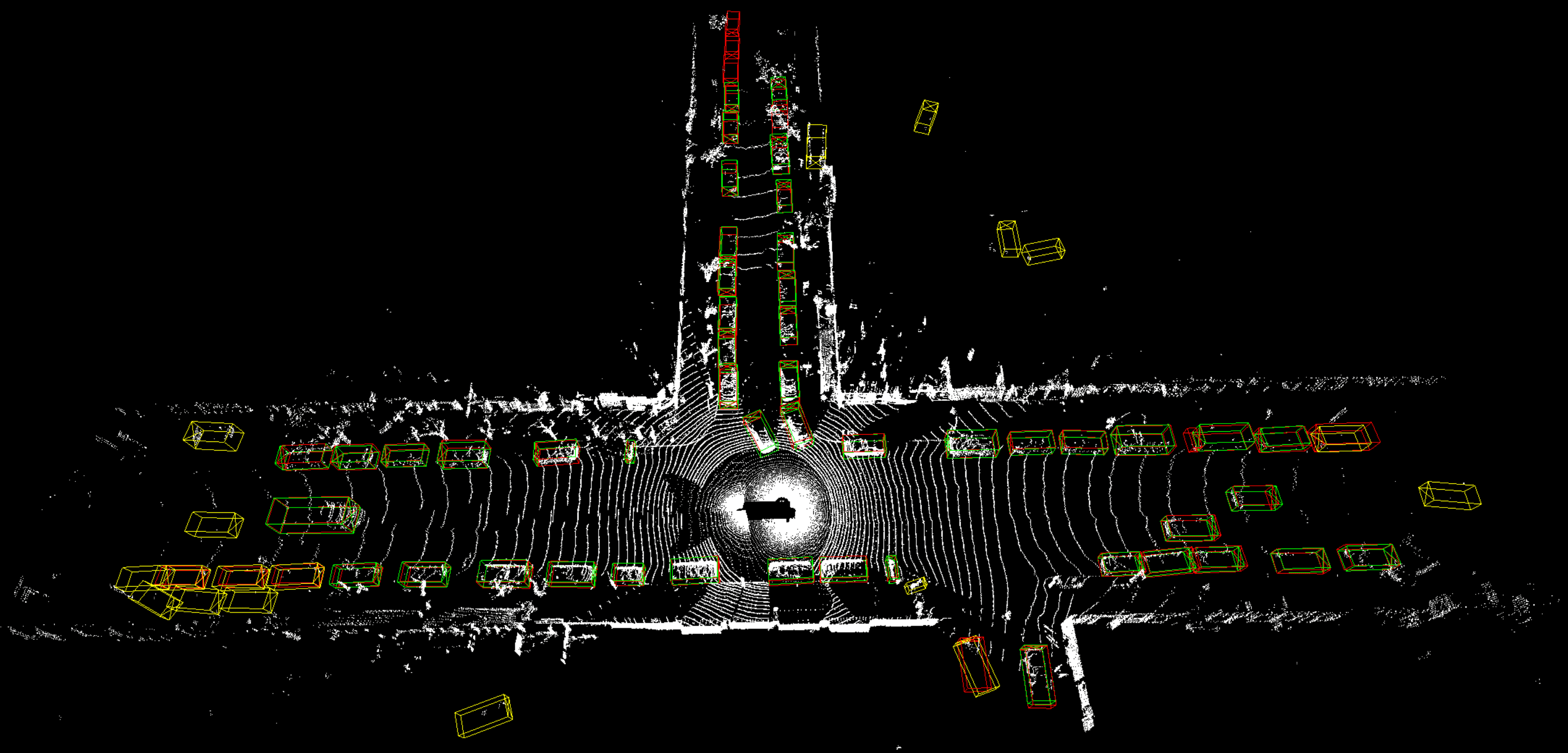}}
		\vspace{1pt}
		\centerline{(b) GLENet-S (Ours)}
	\end{minipage}
	\caption{Visual comparison of the results by SECOND and GLENet-S on the Waymo val set. The ground-truth, true positive and false positive bounding boxes are visualized in red, green and yellow, respectively. Best viewed in color and zoom in for more details. \revise{Additional NMS with a higher IoU threshold is conducted to eliminate overlapped bounding boxes for better visualization.}}
	\label{fig:vis_waymo}
\end{figure*}

%\setlength{\tabcolsep}{1pt}
%\begin{table}[t]
%\centering
%\caption{Inference speed of different detection frameworks on the KITTI dataset.}
%\label{inference}
%\scalebox{0.72}{
%	\small
%	\begin{tabular}{ l| cc | cc | cc}
%		\toprule
%		\thead{Method} & \thead{SECOND \\ \citep{yan2018second}} & \thead{GLENet-S \\ (Ours)} & \thead{CIA-SSD \\ \citep{zheng2021cia}} & \thead{GLENet-C \\ (Ours)} & \thead{Voxel R-CNN \\ \citep{deng2021voxel}} & \thead{GLENet-VR \\ (Ours)} \\
%		\hline
%		FPS(Hz) &     23.36      & 22.80     & 27.18      & 28.76     & 21.08           & 20.82      \\
%		\bottomrule
%\end{tabular}}
%\end{table}
%\setlength{\tabcolsep}{1.5pt}

\setlength{\tabcolsep}{12pt}
\begin{table}
	\centering
	\caption{Inference time comparison for different baselines on the KITTI dataset.}
	\label{inference}
	\begin{tabular}{l|c} 
		\toprule
		Method       & FPS (Hz)  \\ 
		\hline
		SECOND~\citep{yan2018second}      & 23.36     \\
		GLENet-S (Ours)     & 22.80     \\ 
		\hline
		CIA-SSD~\citep{zheng2021cia}     & 27.18    \\
		GLENet-C (Ours)     & 28.76     \\ 
		\hline
		Voxel R-CNN~\citep{deng2021voxel} & 21.08     \\
		GLENet-VR (Ours)    & 20.82    \\
		\bottomrule
	\end{tabular}
\end{table}
\setlength{\tabcolsep}{4pt}

\subsubsection{Inference Efficiency} We evaluated the inference speed of different baselines with a batch size of 1 on a desktop with Intel CPU E5-2560 @ 2.10 GHz and NVIDIA GeForce RTX 2080Ti GPU. As shown in Table~\ref{inference}, our approach does not significantly increase the computational overhead. Particularly, GLENet-VR only takes 0.6 more ms than the base Voxel R-CNN, since the number of candidates for the input of \revise{variance} voting is relatively small in two-stage detectors.

%\subsection{\revise{Visualization of Localization Uncertainty}}

\subsection{Comparison of Visual Results}\label{qualitative_results}
%\bmhead{Visualization of GLENet output}
Fig.~\ref{fig:sm_debug_compare} visualizes the detection results of our GLENet-VR and the baseline Voxel R-CNN on the KITTI val set, where it can be seen that our GLENet-VR obtains better detection results with fewer false-positive bounding boxes and fewer missed heavily occluded and distant objects than Voxel R-CNN. %\JHdel{compared with the base detector}. 
We also compared the detection results of SECOND and GLENet-S on the Waymo validation set in Fig.~\ref{fig:vis_waymo}, where it can be seen that compared with SECOND~\citep{yan2018second}, our GLENet-S has fewer false predictions and achieves more accurate localization.

%\bmhead{Visualization of GLENet output} 
% \JHNOTE{We can merge this paragraph into the caption of Fig. 3} We also include some visualization of results from GLENet. As shown in Fig.~\ref{fig:plenet_out}, given a point cloud object, we can acquire potentially plausible bounding boxes with GLENet by sampling latent variables multiple times. In general, GLENet tends to predict diverse bounding boxes for objects represented with sparse point clouds and incomplete outlines, and consistently accurate boundary boxes for high-quality point cloud objects. Therefore, the variance of GLENet's multiple predictions can represent the label uncertainty in ground-truth bounding boxes.

\section{Discussion}
\noindent
\revise{
%\textbf{Limitations.} While our method addresses the issue of label uncertainty in 3D object detection and achieves promising results, it is important to consider the limitations of your approach. Here are a few potential limitation. 1) 
%\\
%\noindent\textbf{Future Directions.} 
In this section, we further list a few potential technical limitations of the current learning framework and promising directions for extensions.
\begin{enumerate}[label=(\arabic*)]
	\item \textit{Complexity and Computational Cost}. Despite GLENet providing reliable label uncertainty as supervision signals for downstream probabilistic detectors, estimating the label uncertainty itself brings additional computational costs and makes the overall training process more complex. Particularly, considering the risk of over-fitting, we followed k-fold cross-sampling to train GLENet on 9 subsets and then made predictions on the remaining subset at each time.
	\item \textit{Incomplete Input Information}. In GLENet, we only take the partial point cloud of individual objects as input, so only the learned geometric information is used to estimate potential bounding boxes. However, the context cues like free space and location of surrounding objects are neglected, which are also meaningful to determining the bounding boxes. Therefore, the estimated label uncertainty may deviate from the true distribution. But it is not feasible to take all points in the scene as input, as the key point of GLENet lies in learning the latent distribution of bounding boxes from samples with similar point cloud shapes and involving the whole point cloud in the scene distinguishes those objects with similar shapes. Incorporating such information without compromising the core benefits of GLENet remains a challenge.
	\item \textit{Robustness to Annotation Errors}. While GLENet aims to address the inherent ambiguity in ground-truth annotations, it may not be entirely immune to the effects of significant annotation errors.
	If the training data contains substantial annotation errors, the model may inadvertently learn and propagate these errors, leading to an inaccurate estimation of label uncertainty. For example, if an object with a high-quality point cloud is annotated with a wrong box and further leads to inconsistent predictions and larger label uncertainty, those objects with similar shapes will suffer from unreasonable label uncertainty supervision signals.
	The robustness and reliability of the proposed method under such scenarios could be a limitation. 
	\item \textit{Limited Evaluation Metrics and Scenarios}. 
	Evaluating the quality and diversity of generated data in generative tasks like GLENet is challenging. 
%	The difficulty in evaluating the quality of models in generative tasks is a well-worn topic of discussion, as it is often challenging to establish objective criteria and avoid subjectivity in the evaluation process. This issue is also relevant for GLENet, as with other generative models.
	Although the proposed $L_{NLL}$ assesses the closeness between the prediction of GLENet and ground-truth annotation bounding boxes, evaluating the quality and diversity of generated data remains an ongoing research problem.
	On the other hand, while your method demonstrates performance gains on benchmark datasets such as KITTI and Waymo, it is important to consider the generalizability of your approach across various environmental conditions, object classes, and sensor modalities. The ability of GLENet to generalize to a broader range of datasets and scenarios could be a limitation.
	\item \textit{Possible Extensions}. The idea of estimating the label uncertainty by capturing the one-to-many relationship between observed input and multiple plausible labels with latent variables could be extended to other subjective tasks in computer vision where labels are not deterministic. One promising task is 3D object tracking, where different opinions of annotators on the boundaries of objects lead to non-deterministic labels. Another example is image quality assessment, where the goal is to evaluate the quality of an image, often in the context of compression or transmission. The quality of an image is subjective and can vary depending on the perception and expectations of the viewer.
\end{enumerate}
}

\section{Conclusion} 
\label{sec:con}
We presented a general and unified deep learning-based paradigm for modeling 3D object-level label uncertainty. Technically, we proposed GLENet, adapted from the learning framework of CVAE, to capture one-to-many relationships between incomplete point cloud objects and potentially plausible bounding boxes. As a plug-and-play component, GLENet can generate reliable label uncertainty statistics that can be conveniently integrated into various 3D detection pipelines to build powerful probabilistic detectors. We verified the effectiveness and universality of our method by incorporating the proposed GLENet into several existing deep 3D object detectors, which demonstrated consistent improvement and produced state-of-the-art performance on both KITTI and Waymo datasets.

\section*{Data Availability Statements}
The Waymo Open Dataset~\citep{Sun_2020_CVPR} and KITTI~\citep{Geiger_KITTI} used in this manuscript are deposited in publicly available repositories respectively: \url{https://waymo.com/open/data/perception} and \url{http://www.cvlibs.net/datasets/kitti}.

\bibliographystyle{spbasic}
\bibliography{bib}   

\end{document}